\documentclass[11pt,a4paper]{article}
\usepackage[hyperref]{eacl2021}
\usepackage{times}
\usepackage{latexsym}

\usepackage{microtype}

\aclfinalcopy 

\usepackage{multirow}

\usepackage{microtype}
\usepackage{graphicx}
\usepackage{subfigure}
\usepackage{booktabs} 
\usepackage{amsthm}
\usepackage{amsmath}

\usepackage[export]{adjustbox}
\usepackage{float}

\usepackage[outline]{contour}

\usepackage{natbib}

\theoremstyle{definition}
\newtheorem*{definition}{Desiderata}

\theoremstyle{definition}

\theoremstyle{definition}

\newcommand{\argmin}[1]{\underset{#1}{\operatorname{argmin}}\;}

\usepackage{arydshln}

\newtheorem*{defn}{Task Definition}
\usepackage{hyperref}

\definecolor{tablegray}{rgb}{0.2,0.2,0.2}
\newcommand{\stdintable}[1] {~~\textcolor{tablegray}{\scriptsize{$\pm$#1}}}

\definecolor{col1}{RGB}{68,133,62}
\definecolor{col2}{RGB}{202,187,88}
\definecolor{col3}{RGB}{157,62,116}

\newcommand{\greenarrowup}{{\color{col1}\bgroup\contourlength{0.05em}\contour{col1}{$\nearrow $}\egroup}}

\newcommand{\orangearrowright}{{\color{col2}\bgroup\contourlength{0.05em}\contour{col2}{$\rightarrow $}\egroup}}

\newcommand{\redarrowdown}{{\color{col3}\bgroup\contourlength{0.05em}\contour{col3}{$\searrow $}\egroup}}

\title{AdapterFusion: \\ Non-Destructive Task Composition for Transfer Learning}

\author{Jonas Pfeiffer$^{1}$, Aishwarya Kamath$^{2}$, Andreas R\"uckl\'{e}$^1$, \\{\bf Kyunghyun Cho$^{2,3}$, Iryna Gurevych$^{1}$  } \\
$^1$Ubiquitous Knowledge Processing Lab (UKP Lab), 
  Technical University of Darmstadt \\
$^2$New York University \hspace{0.5em} $^3$CIFAR Associate Fellow\\
\texttt{pfeiffer@ukp.tu-darmstadt.de} \\
}

\date{}

\begin{document}
\maketitle
\begin{abstract}

Sequential fine-tuning and multi-task learning are methods aiming to incorporate knowledge from multiple tasks; however, they suffer from catastrophic forgetting and difficulties in dataset balancing. 
To address these shortcomings,
we propose \emph{AdapterFusion}, a new two stage learning algorithm that leverages knowledge from multiple tasks.
First, in the \textit{knowledge extraction} stage we learn task specific parameters called \textit{adapters}, that encapsulate the task-specific information.
We then combine the adapters in a separate \textit{knowledge composition} step. 
We show that by separating the two stages, i.e., knowledge extraction and knowledge composition, the classifier can effectively exploit the representations learned from multiple tasks in a non-destructive manner.
We empirically evaluate 
AdapterFusion
on 16 diverse NLU tasks, and find that it effectively combines various types of knowledge at different layers of the model. We show that our approach
 outperforms traditional strategies such as full  fine-tuning as well as multi-task learning. Our code and
adapters are available at \href{https://AdapterHub.ml}{AdapterHub.ml}. 
\end{abstract}

\section{Introduction}
\label{introduction}

\begin{figure}[htp]
\centering
\includegraphics[width=0.4\linewidth]{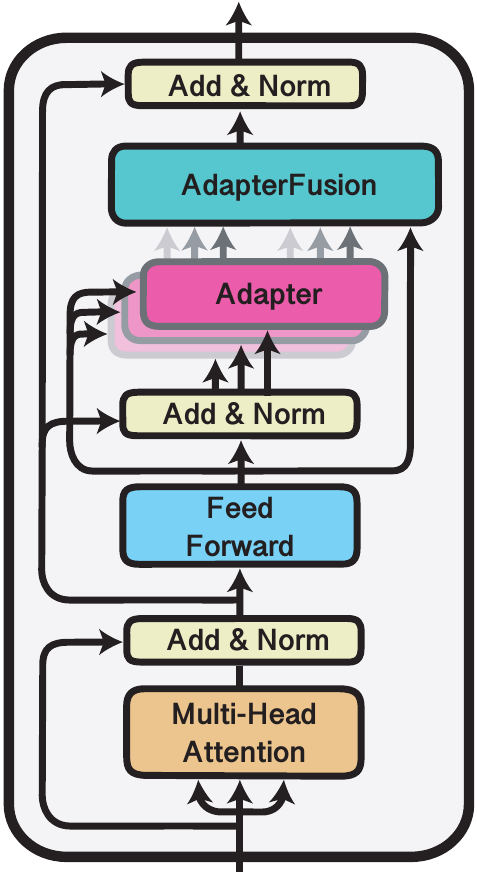}
\caption{AdapterFusion architecture inside a  transformer \cite{Vaswani17}. The AdapterFusion component takes as input the representations of multiple adapters trained on different tasks and learns a parameterized mixer of the encoded information.}
\label{fig:Adapter_Fusion_high_level}
\end{figure}

The most commonly used method
for solving NLU tasks is to leverage pretrained models, with the dominant architecture being a transformer \cite{Vaswani17},  typically trained with a language modelling objective \cite{devlin2018bert, radford2018improving, liu2019roberta}. 
Transfer to a task of interest is achieved by fine-tuning all the weights of the pretrained model on that \emph{single task}, often yielding state-of-the-art results  \cite{zhang2017survey, ruder2017, howard2018universal, peters2019tune}. 
However, each task of interest requires all the parameters of the network to be fine-tuned,
which results in a specialized model for each task.

There are two approaches for sharing information across multiple tasks. The first consists of starting from the pretrained language model and sequentially fine-tuning on each of the tasks one by one \cite{phang2018sentence}. 
However, as we subsequently fine-tune the model weights on new tasks, the problem of catastrophic forgetting 
\cite{mccloskey1989catastrophic,french1999catastrophic} can arise, which results in loss of knowledge already learned 
from all previous tasks.
This, together with the non-trivial decision of the order of tasks in which to fine-tune the model, 
hinders the effective transfer of knowledge. 
Multi-task learning \cite{caruana1997multitask, zhang2017survey, liu2019multi} is another approach
for sharing information across multiple tasks. 
This involves fine-tuning the weights of a pretrained language model using a weighted sum of the objective function of each target task simultaneously. Using this approach, 
the network captures the common structure underlying all the target tasks. However, multi-task learning requires simultaneous access to all tasks during training. Adding new tasks thus requires complete joint retraining. 
Further, it is 
difficult
to balance multiple tasks and train a model that solves each task equally well. As has been shown in \citet{lee2017fully}, these models often overfit on low resource tasks and underfit on high resource tasks. 
This makes it difficult to effectively transfer knowledge across tasks with all the tasks being solved equally well \cite{pfeiffer20multilabel}, thus considerably limiting the applicability of multi-task learning in many scenarios.

Recently, \emph{adapters} \cite{rebuffi2017learning, houlsby2019parameter} have emerged as an alternative training strategy. 
Adapters do not require fine-tuning of
all parameters of the pretrained 
model, and instead introduce a small number of task specific parameters --- 
while keeping the underlying pretrained language model fixed.
Thus, we can separately and simultaneously train adapters for multiple tasks, which all share the same underlying pretrained parameters. 
However, to date, there exists no method for using \emph{multiple} adapters to maximize the transfer of knowledge across tasks without suffering from the same problems as sequential fine-tuning and multi-task learning. For instance,
\citet{stickland2019bert} propose a multi-task approach for training adapters, which still suffers from the difficulty of balancing the various target tasks and requiring simultaneous access to all target tasks.
 
In this paper we address these limitations and propose a new variant of adapters 
called \emph{AdapterFusion}. We further propose a novel two stage learning algorithm that allows us to effectively share knowledge across multiple tasks while avoiding the issues of catastrophic forgetting and balancing of different tasks. Our AdapterFusion architecture, illustrated in Figure \ref{fig:Adapter_Fusion_high_level}, has two components. The first component is an adapter trained on a task without changing the weights of the underlying language model. The second component --- our novel Fusion layer --- combines
the representations from several such task adapters in order to improve the performance on the target task.

\paragraph{Contributions} Our main contributions are:
(1) We introduce a novel two-stage transfer learning strategy, termed \textit{AdapterFusion}, which combines the knowledge from multiple source tasks to perform better on a target task. 
(2) We empirically evaluate our proposed approach on a set of 16 diverse NLU tasks such as sentiment analysis, commonsense reasoning, paraphrase detection, and recognizing textual entailment. 
(3) We compare our approach with 
\citet{stickland2019bert} where  adapters are trained for all tasks in a multi-task manner, finding that AdapterFusion is able to improve this method, even though the model has simultaneous access to all tasks during pretraining. 
(4)~We show that our proposed approach outperforms fully fine-tuning the transformer model on a single target task. Our approach additionally outperforms adapter based models trained both in a Single-Task, as well as Multi-Task setup. 

The code of this work is integrated into the \href{https://AdapterHub.ml}{AdapterHub.ml} \cite{pfeiffer2020AdapterHub}.

\section{Background}
\label{task}

In this section, we formalize 
our goal of transfer learning \cite{pan2009survey, torrey2010transfer, ruder2019neural},
highlight its key challenges, and provide a brief overview of common methods 
that can be used to address them. This is followed by an introduction to \emph{adapters} \cite{rebuffi2017learning} and a brief formalism of the two approaches to training adapters.

\begin{defn}
We are given a model that is pretrained on a task with training data $D_0$ and a loss function $L_0$. 
The weights $\Theta_0$ of this model are learned as follows:
\begin{align*}
    D_0&:=\textrm{Large corpus of unlabelled text} \\
    L_0&:= \textrm{Masked language modelling loss} \\
    \Theta_0 &\leftarrow \argmin{\Theta}L_0(D_0;\Theta) 
\end{align*}
In the remainder of this paper, we refer to this pretrained model by the tuple $( D_0, L_0)$.

We define $C$ as the set of N classification tasks having labelled data of varying sizes and different loss functions: 
\begin{gather*}
    C = \{ ( D_1, L_1), \ldots ,(D_N, L_N)\}
\end{gather*}

The aim is to be able to leverage a set of $N$ tasks to improve on a target task $m$ with $C_m = { ( D_m, L_m)}$. 
In this work we  focus on the setting where $m \in \{1,\ldots, N\}$. 
\end{defn}

\begin{definition}
We wish to learn a parameterization $\Theta_m$ that is defined as follows:
\begin{gather*}
    \Theta_m \leftarrow \argmin{\Theta'}L_m(D_m;\Theta') 
\end{gather*}
where $\Theta'$ is expected to have encapsulated relevant information from all the $N$ tasks. The target model for task $m$ is initialized with $\Theta'$ for which we learn the optimal parameters  $\Theta_m$  through minimizing the task's loss on its training data.  
\end{definition}

\subsection{Current Approaches to Transfer Learning}
\label{sec:Current_approaches}

There are two predominant approaches to achieve sharing of information from one task to another.

\subsubsection{Sequential Fine-Tuning} This involves sequentially updating all the weights of the model on each task.  
For a set of $N$ tasks, the order of fine-tuning is defined and at each step  the model is initialized with the parameters learned through the previous step.
However, this approach does not perform well beyond two sequential tasks \cite{phang2018sentence, pruksachatkun2020intermediate} due to catastrophic forgetting.

\subsubsection{Multi-Task Learning (MTL)} All tasks are trained simultaneously with the aim of learning a shared representation that will enable the model to generalize better on each task \cite[][\textit{inter alia}]{caruana1997multitask, collobert2008unified, nam2014large, liu2016recurrent,  liu2017adversarial, zhang2017survey, ruder2017, ruder2019latent, sanh2019hierarchical, pfeiffer20multilabel}.
\begin{equation*}
\begin{split}
    \Theta_{0 \rightarrow \{1, ..., N\}} \leftarrow \argmin{\Theta} \left( \sum_{n=1}^N L_n(D_n;\Theta_0)  \right)
\end{split}
\end{equation*}
Where $\Theta_{0 \rightarrow \{1, ..., N\}}$ indicates that we start with $\Theta_{0}$ and fine-tune on a set of tasks $\{1, ..., N\}$.

 However, MTL requires simultaneous access to all tasks, making it difficult to add more tasks on the fly. As the different tasks have varying sizes as well as loss functions, effectively combining them during training is very challenging and requires heuristic approaches as proposed in \citet{stickland2019bert}.

\subsection{Adapters}
\label{sec:Adapter_Chapter}

While 
the predominant
methodology for transfer learning 
is 
to fine-tune all weights of the pretrained model,
\emph{adapters}  \cite{houlsby2019parameter}
have 
recently
been introduced as an alternative 
approach
with applications in domain transfer \cite{rueckle2020MultiCQA}, machine translation \cite{bapna2019simple, Philip20NaverAdapter} transfer learning \cite{stickland2019bert, Wang20kadapters, Lauscher2020comonsense}, and cross-lingual transfer  \cite{ pfeiffer20madx, Pfeiffer20Gib, ustun-etal-2020-udapter, Vidoni2020OrthogonalLA}. 
Adapters 
share a large set of parameters $\Theta$  across all tasks and introduce a small number of task-specific parameters $\Phi_n$.
While $\Theta$ represents the weights of a pretrained model (e.g., a transformer), the parameters  $\Phi_n$, where $n \in \{1,\ldots, N\}$, are used to encode task-specific representations in intermediate layers of the shared model. 
\label{sec:conceptual}
Current work on adapters focuses either on training adapters for each task separately  \cite{houlsby2019parameter,bapna2019simple,pfeiffer2020AdapterHub} 
or training them in a multi-task setting to leverage shared representations \cite{stickland2019bert}. 
We discuss both variants below.

\subsubsection{Single-Task Adapters (ST-A)}
For each of the $N$ tasks, the model is initialized with parameters $\Theta_0$.
In addition, a set of new and randomly initialized adapter parameters  $\Phi_n$ are introduced. 

The parameters $\Theta_0$ are fixed and only the parameters $\Phi_n$ are trained. 
This makes it possible to efficiently parallelize the training of adapters for all $N$ tasks, and store the corresponding knowledge in designated parts of the model. 
The objective for each task $n \in \{1,\ldots, N\}$ is of the form:
\begin{align*}
    \Phi_n &\leftarrow \argmin{\Phi}L_n(D_n;\Theta_0,\Phi)
\end{align*}
For common adapter architectures, $\Phi$ contains considerably fewer parameters than $\Theta$, e.g.,
only 3.6\% of the parameters of the pretrained model in \citet{houlsby2019parameter}.

\subsubsection{Multi-Task Adapters (MT-A)} 
\citet{stickland2019bert} propose to train adapters for $N$ tasks in parallel with a multi-task objective. The underlying parameters
$\Theta_0$
are fine-tuned along with the task-specific parameters in $\Phi_n$.
The training objective can be defined as:

\begin{equation*}
\begin{split}
     \mathbf{\Theta} \leftarrow 
    \argmin{\Theta, \Phi} \left( \sum_{n=1}^N L_n(D_n;\Theta_0,\Phi_n)\right)
\end{split}
\end{equation*}

where
\begin{gather*}
    \mathbf{\Theta} =  \Theta_{0 \rightarrow \{1, ..., N\}},  
    \Phi_1, \dots, \Phi_N.
\end{gather*}

\subsubsection{Adapters in Practice}
\label{sec:adapterinpractice}
Introducing new adapter parameters in different layers of an otherwise fixed pretrained model has been shown to perform on-par with, or only slightly below, full model fine-tuning 
\cite{houlsby2019parameter, stickland2019bert, pfeiffer2020AdapterHub}.
For NLP tasks, adapters have been introduced for the transformer architecture \cite{Vaswani17}. At each transformer layer $l$, a set of adapter parameters $\Phi_l$ is introduced. 
The placement and architecture of adapter parameters $\Phi$ within a pretrained model is non-trivial.  \citet{houlsby2019parameter} experiment with different architectures, finding that a two-layer feed-foward neural network with a bottleneck works well. They place two of these components within one layer, one after the multi-head attention (further referred to as \textit{bottom}) and one after the feed-forward layers of the transformer (further referred to as \textit{top}).\footnote{ We illustrate these placements in Appendix Figure~\ref{fig:Adapter_Architecture} (left).}
\citet{bapna2019simple} and \citet{stickland2019bert} only introduce one of these components at the \textit{top} position, however, \citet{bapna2019simple}  include an additional \textit{layer norm} \cite{Ba16_LayerNorm}.

Adapters trained in both single-task (ST-A) or multi-task (MT-A) setups have learned the idiosyncratic knowledge of the respective tasks' training data, encapsulated in their  designated parameters.  
This results in a compression of information, which requires less space 
to store task-specific knowledge. 
However, the distinct weights of adapters prevent a downstream task from being able to use multiple sources of extracted information.
In the next section we describe our two stage algorithm which tackles the sharing of information stored in adapters trained on different tasks.

\section{AdapterFusion}
\label{fusion}

Adapters avoid catastrophic forgetting by introducing task-specific parameters; however, current adapter approaches do not allow sharing of information between tasks. 
To mitigate this we propose AdapterFusion.
 
\subsection{Learning algorithm}
\label{sec:fusion:learning}
In the first stage of our learning algorithm, we train either ST-A or MT-A for 
each of the N tasks. 
 
In the second stage, we then combine the set of $N$ adapters by using AdapterFusion.  
While fixing both the parameters $\Theta$ as well as all adapters $\Phi$, we introduce parameters $\Psi$ that learn to combine the $N$ task adapters to solve the target task. 
\begin{gather*}
    \Psi_m \leftarrow \argmin{\Psi}L_m(D_m;\Theta,\Phi_1,\ldots,\Phi_N, \Psi) 
\end{gather*}
$\Psi_m$ are the newly learned AdapterFusion parameters for task $m$. $\Theta$ refers to $\Theta_0$ in the ST-A setting or $\Theta_{0 \rightarrow \{1, ..., N,m\}}$ in the MT-A setup.  In our experiments we focus on the setting where $m \in \{1,...,N\}$, which means that the training dataset of $m$ is used twice: \textit{once} for training the adapters $\Phi_m$ and \textit{again} for training Fusion  parameters $\Psi_m$, which learn to compose the information stored in the $N$ task adapters. 

By separating the two stages --- knowledge extraction in the adapters, and knowledge composition with AdapterFusion --- we address the issues of catastrophic forgetting, interference between tasks and training instabilities.

\begin{figure}
\centering
\includegraphics[width=0.6\linewidth]{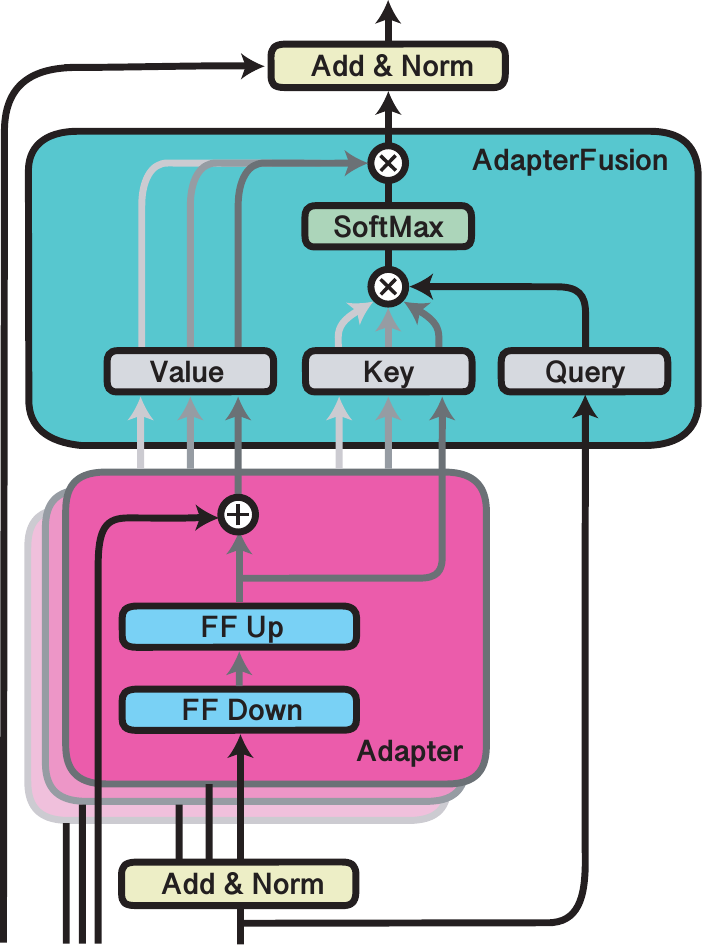}
\caption{Our AdapterFusion architecture. This includes learnable weights \textit{Query}, \textit{Key}, and \textit{Value}. \textit{Query} takes as input the output of the pretrained transformer weights. Both \textit{Key} and \textit{Value} take as input the output of the respective adapters. The dot product of the \textit{query} with all the \textit{keys} is passed into a softmax function, which learns to weight the adapters with respect to the context. }
\label{fig:AdapterFusion}
\end{figure}

\subsection{Components}
\label{sec:fusion:components}
AdapterFusion learns to 
compose the 
$N$ task adapters $\Phi_n$ and the shared pretrained model $\Theta$, by introducing a new set of weights $\Psi$.
These parameters learn to combine the adapters as a dynamic function of the target task data. 

As illustrated in Figure \ref{fig:AdapterFusion}, we define the AdapterFusion parameters $\Psi$ to consist of \textit{Key, Value} and \textit{Query} matrices at each layer $l$, denoted by $\textbf{K}_l$, $\textbf{V}_l$ and $\textbf{Q}_l$ respectively. 
At \emph{each layer} $l$ of the transformer and  \emph{each time-step} $t$, the output of the feed-forward sub-layer of layer $l$ is taken as the query vector. 
The output of each adapter $\textbf{z}_{l,t}$ is used as input to both the \textit{value} and \textit{key} transformations. 
Similar to attention \cite{Bahdanau14Attention, Vaswani17}, we learn a contextual activation of each adapter $n$ using
\begin{alignat*}{2}
    &\textbf{s}_{l,t} &&= \text{softmax}(\textbf{h}_{l,t}^\top\textbf{Q}_l \otimes \textbf{z}_{l,t,n}^\top\textbf{K}_l),  n\in\{1,{\scriptscriptstyle \ldots}, N\} 
\\
    &\textbf{z}_{l,t,n}' &&= \textbf{z}_{l,t,n}^\top\textbf{V}_l,  n\in\{1,{\scriptscriptstyle \ldots}, N\}
\\
    &\textbf{Z}_{l,t}' &&= [\textbf{z}_{l,t,0}', {\scriptscriptstyle \ldots} ,\textbf{z}_{l,t,N}']
\\
    &\textbf{o}_{l,t} &&= \textbf{s}_{l,t}^\top \textbf{Z}_{l,t}'
\end{alignat*}
Where $\otimes$ represents the dot product and $[\cdot, \cdot]$ indicates the concatenation of vectors.

Given the context, AdapterFusion learns a parameterized mixer of the available trained adapters. It learns to identify and activate the most useful adapter for a given input.

\begin{table*}[h]
\centering
\footnotesize
\begin{tabular}{l|ll:ll:ll|l}
\toprule
\textbf{Dataset} &   \textbf{Head} &   \textbf{Full} & \textbf{ST-A}  & \textbf{MT-A}& \textbf{F. w/ ST-A} & \textbf{F. w/ MT-A}  & \textbf{ST-A\textsuperscript{Houlsby}}\\  
\midrule 
MNLI	&	54.59				    &	84.10				    &	\textbf{84.32}  		&	82.49		\stdintable{0.49}	&	84.28		    		&	83.05				&	84.13		\\
QQP	        &	76.79				    &	\textbf{90.87}	    	&	90.59	        		&	89.47		\stdintable{0.60}	&	90.71		    		&	90.58				&	90.63		\\
SST     	&	85.17		\stdintable{0.45}   &	92.39		\stdintable{0.22}	&	91.85		\stdintable{0.41}	&	92.27		\stdintable{0.71}	&	92.20		\stdintable{0.18}	&	\textbf{93.00}		\stdintable{0.20}	&	92.75\stdintable{0.37}	\\
WGrande	&	51.92		\stdintable{0.35}   &	60.01		\stdintable{0.08}	&\textbf{61.09} \stdintable{0.11}    &	57.70		\stdintable{1.40}	&	60.23		\stdintable{0.31}	&	59.32		\stdintable{0.30}	&	59.32\stdintable{1.33}	\\
IMDB    	&	85.05		\stdintable{0.22}   &\textbf{94.05}	\stdintable{0.21}	&	93.85		\stdintable{0.07}	&	92.56		\stdintable{0.54}	&	93.82		\stdintable{0.39}	&	92.66		\stdintable{0.32}	&	93.96\stdintable{0.22}	\\
HSwag	&	34.17		\stdintable{0.27}   &\textbf{39.25}	\stdintable{0.76}	&	38.11		\stdintable{0.14}	&	36.47		\stdintable{0.98}	&	37.98	    \stdintable{0.01}	&	37.36 \stdintable{0.10}			&	38.65\stdintable{0.25}	\\
\hdashline
SocIQA	&	50.33		\stdintable{2.50}   &	62.05		\stdintable{0.04}	&	62.41		\stdintable{0.11}	&	61.21		\stdintable{0.89}	&\textbf{63.16}	\stdintable{0.24}	&	62.56		\stdintable{0.10}			&	62.73\stdintable{0.53}	\\
CosQA	&	50.06		\stdintable{0.51}	&	60.28		\stdintable{0.40}   &	60.01		\stdintable{0.02}	&	61.25		\stdintable{0.90}	&	60.65		\stdintable{0.55}	&	\textbf{62.78}		\stdintable{0.07}	&	61.37\stdintable{0.35}	\\
SciTail	    &	85.30		\stdintable{2.44}   &	94.32		\stdintable{0.11}	&	93.90		\stdintable{0.16}	&	94.53		\stdintable{0.43}	&	94.04		\stdintable{0.23}	&	\textbf{94.79}		\stdintable{0.17}	&	94.07\stdintable{0.39}	\\
Argument	&	70.61		\stdintable{0.59}	&	76.87		\stdintable{0.32}	&\textbf{77.65}	\stdintable{0.34}	&	75.70		\stdintable{0.60}	&\textbf{77.65}\stdintable{0.21}	&	76.08		\stdintable{0.27}	&	77.44\stdintable{0.62}	\\
\hdashline
CSQA    	&	41.09		\stdintable{0.27}	&	58.88		\stdintable{0.40}	&	58.91		\stdintable{0.57}	&	53.30		\stdintable{2.19}	&59.73	\stdintable{0.54}	&	56.73		\stdintable{0.14}	&	\textbf{60.05}\stdintable{0.36}	\\
BoolQ	    &	63.07		\stdintable{1.27}	&	74.84		\stdintable{0.24}	&	75.66		\stdintable{1.25}	&	78.76		\stdintable{0.76}	&	76.25		\stdintable{0.19}	&	\textbf{79.18} \stdintable{0.45} &	76.02\stdintable{1.13}	\\
MRPC    	&	71.91		\stdintable{0.13}	&	85.14		\stdintable{0.45}	&	85.16		\stdintable{0.52}	&	81.86		\stdintable{0.99}	&\textbf{90.29}	\stdintable{0.84}	&	84.68		\stdintable{0.32}	&	86.66\stdintable{0.81}	\\
\hdashline
SICK	    &	76.30		\stdintable{0.71}   &	87.30		\stdintable{0.42}	&	86.20		\stdintable{0.00}	&	88.61		\stdintable{1.06}	&	87.28		\stdintable{0.99}	&	\textbf{90.43}		\stdintable{0.30}	&	86.12\stdintable{0.54}	\\
RTE	        &	61.37		\stdintable{1.17}	&	65.41		\stdintable{0.90}	&	71.04		\stdintable{1.62}	&	77.61		\stdintable{3.21}	&	76.82		\stdintable{1.68}	&	\textbf{79.96}		\stdintable{0.76}	&	69.67\stdintable{1.96}	\\
CB	        &	68.93		\stdintable{4.82}	&	82.49		\stdintable{2.33}	&	86.07		\stdintable{3.87}	&	89.09		\stdintable{1.15}	&\textbf{92.14}	\stdintable{0.97}	&	89.81		\stdintable{0.99}	&	87.50\stdintable{4.72}	\\

\midrule
\textbf{Mean}	&	64.17	  	& 75.51 				&  76.05	 	& 	 75.80		 	& \textbf{77.33}	 	 	& 77.06	  & 76.32	\\
\bottomrule
\end{tabular}
\caption{ Mean and standard deviation results (development sets) for each of the 16 datasets and the different architectural setups. The datasets are ordered by their respective training dataset size. Dashed horizontal lines separate datasizes \{$>40k, >10k, >5k$\}, respectively. Each model is initialized with BERT-base \cite{devlin2018bert} weights. \textbf{Head} indicates training only a classification head on top of fixed BERT weights. For \textbf{Full} training we fine-tune all weights of BERT. Single-Task Adapters (\textbf{ST-A}) is the training of independently trained adapters for each task, using the architecture illustrated in Figure \ref{fig:Adapter_Architecture}. Multi-Task Adapters (\textbf{MT-A}) shows results of jointly trained adapters using the default settings of \citet{stickland2019bert}. \textbf{Fusion w/ ST-A} and \textbf{Fusion w/ MT-A} show the results of AdapterFusion using the respective pre-trained Adapters.  
\textbf{ST-A\textsuperscript{Houlsby}}
shows the results of ST-Adapters with the architecture proposed by \citet{houlsby2019parameter}. Reported results are accuracy scores.
}
\label{table:results}
\end{table*}
\section{Experiments}
\label{exp}

In this section
we evaluate how effective AdapterFusion is in overcoming the issues faced by other transfer learning methods. We provide a brief description of the 16 diverse datasets that we use for our study, each of which uses accuracy as the scoring metric. 

\subsection{Experimental Setup}
\label{sec:setup}

In order to investigate our model's ability to overcome catastrophic forgetting, we compare Fusion using ST-A to only the ST-A for the task. We also compare Fusion using ST-A to MT-A for the task to test whether our two-stage procedure alleviates the problems of interference between tasks. Finally, our experiments to compare MT-A with and without Fusion let us investigate the versatility of our approach. Gains in this setting would show that AdapterFusion is useful even when the base adapters have already been trained jointly.

In all experiments, we use BERT-base-uncased \cite{devlin2018bert} as the pretrained language model. 
We train 
ST-A, described in Appendix~\ref{sec:adapter_design} and illustrated in Figure \ref{fig:Adapter_Architecture}, for all datasets described in \S \ref{sec:datasets}. We train them with reduction factors\footnote{A reduction factor indicates the factor by which the hidden size is reduced such that the bottle-neck size for BERT Base with factor 64 is reduced to 12 ($768/64=12$). } $\{2,16,64\}$ and learning rate 0.0001 with AdamW and a linear learning rate decay. We train for a maximum of 30 epochs with early stopping. 
We follow the setup used in \citet{stickland2019bert} for training the
MT-A. We use the default hyperparameters\footnote{We additionally test out batch sizes $16$ and $32$.}, and train a 
MT-A model on all datasets simultaneously.

For AdapterFusion, we empirically find that a learning rate of $5e-5$ works well, and use this in all experiments.\footnote{We have experimented with learning rates $\{6e-6$, $ 5e-5$, $1e-4$, $2e-4\}$} We train for a maximum of 10 epochs with early stopping. 
While we initialize $\textbf{Q}$ and $\textbf{K}$ randomly, we initialize $\textbf{V}$ with a diagonal of ones and the rest of the matrix with random weights having a small norm ($1e-6$). Multiplying the adapter output with this value matrix $\textbf{V}$  initially adds small amounts of noise, but retains the overall representation. We continue to regularize the \textit{Value} matrix using $l_2$-norm to avoid introducing additional capacity.

\subsection{Tasks and Datasets}
\label{sec:datasets} 
We briefly summarize the different types of \emph{tasks} that we include in our experiments, and reference the related datasets accordingly. A detailed descriptions can be found in Appendix~\ref{sec:datasets_app}.  

\textbf{Commonsense reasoning} is used to gauge whether the model can perform basic reasoning skills: 
\textit{Hellaswag}~\cite[][]{zellers18swag, zellers19hellaswag}, \textit{Winogrande}~\cite[][]{Sakaguchi19winogrande}, \textit{CosmosQA}~ \cite[][]{Huang19cosmosqa}, \textit{CSQA}~\cite[][]{Talmor19CSQA}, \textit{SocialIQA}~\cite[][]{Sap19SocialIQA}. \textbf{Sentiment analysis} predicts whether a given text
has a
positive or negative sentiment: \textit{IMDb}~\cite[][]{Maas11IMDB}, \textit{SST}~\cite[][]{socher-etal-2013-recursive}. 
\textbf{Natural language inference} predicts whether one sentence entails, contradicts, or is neutral to another: \textit{MNLI}~\cite[][]{Williams18MNLI}, \textit{SciTail}~\cite[][]{ Khot18SciTail}, \textit{SICK}~\cite[][]{ marelli-etal-2014-sick}, \textit{RTE}~(as combined by \citet[][]{wang18glue}), \textit{CB}~\cite[][]{deMarneffe2019commitmentbank}. \textbf{Sentence relatedness} captures whether two sentences include similar content: \textit{MRPC}~\cite[][]{Dolan05MRPC}, \textit{QQP}\footnote{\href{http://data.quora.com/First-Quora-DatasetReleaseQuestion-Pairs}{data.quora.com/First-Quora-DatasetReleaseQuestion-Pairs}}. We also use an argument mining \textit{Argument}~\cite[][]{Stab18Argument} and reading comprehension \textit{BoolQ}~\cite[][]{Clark19BoolQ} dataset.

\section{Results}
\label{results}

We present results for all 16 datasets 
in Table~\ref{table:results}. 
For reference, we also include the adapter architecture of~\citet{houlsby2019parameter}, ST-A\textsuperscript{Houlsby}, which has twice as many parameters compared to \mbox{ST-A}. 
To provide a fair comparison to~\citet{stickland2019bert} we primarily experiment with BERT-base-uncased. We additionally validate our best model configurations --- \mbox{ST-A} and Fusion with \mbox{ST-A} --- with RoBERTa-base, for which we present our results in Appendix Table~\ref{table:results_roberta}.

\subsection{Adapters}

Training only a prediction-head on the output of a pretrained model can also be considered an adapter. This procedure, commonly referred to as training only the \textit{Head},  performs considerably worse than fine-tuning all weights \cite{howard2018universal, peters2019tune}. We show that the performance of only fine-tuning the \textit{Head} compared to \textit{Full} fine-tuning causes on average a drop of 10 points in accuracy. This demonstrates the need for more complex adaptation approaches. 

In Table~\ref{table:results} we show the results for MT-A and ST-A with a reduction factor $16$ (see the appendix Table \ref{table:results_st_bert} for more results) which we find has a good trade-off between the number of newly introduced parameters and the task performance. Interestingly, the ST-A have a regularization effect on some datasets, resulting in better performance on average for certain tasks, even though a much small proportion of weights is trained. On average, we improve $0.66\%$ by training ST-A instead of the \textit{Full} model.

For MT-A we find that there are considerable performance drops of more than $2\%$ for \textit{CSQA} and \textit{MRPC}, despite the heuristic strategies for sampling from the different datasets \citep{stickland2019bert}. This indicates that these heuristics only partially address common problems of multi-task learning such as catastrophic interference. It also shows that learning a shared representation jointly 
does not guarantee the best results for all tasks.  On average, however, we do see a performance increase of $0.4\%$ using MT-A 
over
\textit{Full} fine-tuning on each task separately, 
which demonstrates
that there are advantages in leveraging information from other tasks with multi-task learning. 

\subsection{AdapterFusion}
\label{sec:results:fusion}
AdapterFusion aims to improve performance on a given target task $m$ by transferring task specific knowledge from the set of all $N$ task adapters, where $m \in \{$1, \ldots, N$\}$.
We hypothesize that if there exists at least one task that supports the target task,  AdapterFusion should lead to performance gains. If no such task exists, then the performance should remain the same.

\newcommand{\STAB}[1]{\begin{tabular}{@{}c@{}}#1\end{tabular}}

\begin{figure*}
\centering
\includegraphics[width=\linewidth]{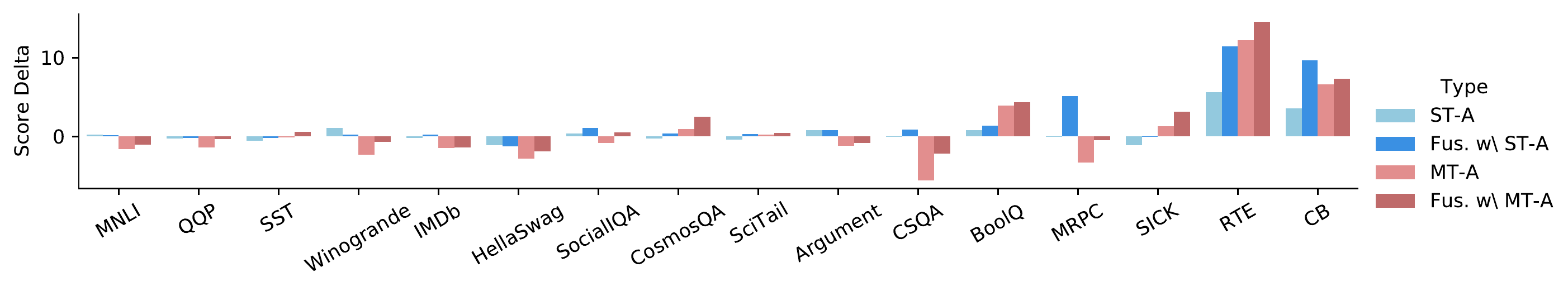}
\caption{Relative performance difference of the two adapter architectures and the AdapterFusion models over fully fine-tuned BERT. Fusion improves over its corresponding adapters (ST-A and MT-A) for most tasks.  }
\label{fig:delta_fusion}
\end{figure*}

\paragraph{Dependence on the size of training data.} 
In Table \ref{table:results} we notice that having access to relevant tasks considerably improves the performance for the target task when using AdapterFusion. While datasets with more than 40k training instances  perform well without Fusion, smaller datasets with fewer training instances benefit more from our approach. We observe particularly large performance gains for datasets with less than 5k training instances. 
For example, Fusion with ST-A achieves substantial improvements of $6.5$ \% for \textit{RTE} and $5.64$ \% for \textit{MRPC}. In addition, we also see performance gains for moderately sized datasets such as the commonsense tasks \textit{CosmosQA} and \textit{CSQA}. Fusion with MT-A achieves smaller improvements, as the model already includes a shared set of parameters. However, we do see performance gains for \textit{SICK}, \textit{SocialIQA}, \textit{Winogrande} and \textit{MRPC}.   
On average, we observe improvements of $1.27$\% and $1.25$\% when using Fusion with ST-A and MT-A, respectively.

\begin{table}[]
\centering
\footnotesize
\begin{tabular}{lcccc}
\toprule
  &  \multicolumn{2}{c}{\textbf{ Fus. w/ ST-A}}   & \multicolumn{2}{c}{\textbf{ Fus. w/ MT-A}} \\
 compared  to   &  ST-A  & MT-A     & ST-A & MT-A    \\
  \midrule
 MNLI	&  \orangearrowright  &  \greenarrowup  &   \redarrowdown  &  \greenarrowup  	\\
QQP			&  \orangearrowright  &  \greenarrowup  &  \orangearrowright  &  \greenarrowup   	\\ 
 SST			&  \greenarrowup  &  \orangearrowright  &   \greenarrowup  &  \greenarrowup   \\
Winogrande	&  \redarrowdown  &  \greenarrowup  &   	 \redarrowdown  &  \greenarrowup   				\\ 
  IMDB		&  \greenarrowup  &  \greenarrowup  &   \redarrowdown  &  \orangearrowright   	\\
HellaSwag	&  \orangearrowright  &  \greenarrowup  &  \redarrowdown  &  \greenarrowup   				\\
\hdashline
  SocialIQA	&  \greenarrowup  &  \greenarrowup   &   \orangearrowright  &  \greenarrowup   	\\
  CosmosQA	& 	 \greenarrowup  &  \redarrowdown  &   \greenarrowup  &  \greenarrowup  \\
  SciTail		&  \orangearrowright  &  \greenarrowup  &   \greenarrowup  &  \orangearrowright   	\\
Argument	&  \orangearrowright  &  \greenarrowup  &   \redarrowdown  &  \greenarrowup  							\\
\hdashline
CSQA		&  \greenarrowup  &  \greenarrowup  &   \redarrowdown  &  \greenarrowup  	\\
BoolQ		& 	  \greenarrowup  &  \redarrowdown  &  \greenarrowup  &  \greenarrowup  	\\
MRPC		&  \greenarrowup  &  \greenarrowup  &   \redarrowdown  &  \greenarrowup  	\\
\hdashline
SICK		&  \greenarrowup  &  \redarrowdown  &   \greenarrowup  &  \greenarrowup   	\\
RTE			&  \greenarrowup  &  \redarrowdown  &   \greenarrowup  &  \greenarrowup  	\\
CB			&  \greenarrowup  &  \greenarrowup  &   \greenarrowup  &  \greenarrowup   \\

\midrule
Improved    &   10/16       &   11/16       &       7/16        &   14/16   \\
\bottomrule
\end{tabular}
\caption{Performance changes of AdapterFusion compared to ST-A and MT-A. Arrows indicate whether there has been an improvement \greenarrowup  \,  ($>0.3$), decrease \redarrowdown  \,  ($<-0.3$), or whether the results have stayed the same \orangearrowright  \,  $[-0.3,0.3]$. 
}
\label{table:AF_arrows}
\end{table}

\paragraph{Mitigating catastrophic interference.}
In order to identify whether our approach is able to mitigate problems faced by multi-task learning, we present 
the performance differences of adapters and AdapterFusion compared to the fully fine-tuned model
in Figure~\ref{fig:delta_fusion}.
In Table~\ref{table:AF_arrows}, we compare AdapterFusion to ST-A and MT-A. The arrows indicate whether there is an improvement \greenarrowup, decrease \redarrowdown, or if the the results remain the same \orangearrowright. We compare the performance of  both, Fusion with ST-A and Fusion with MT-A, to ST-A and MT-A. We summarize our four most important findings below.

\begin{figure*}[ht!]
\centering
\includegraphics[width=1.0\linewidth]{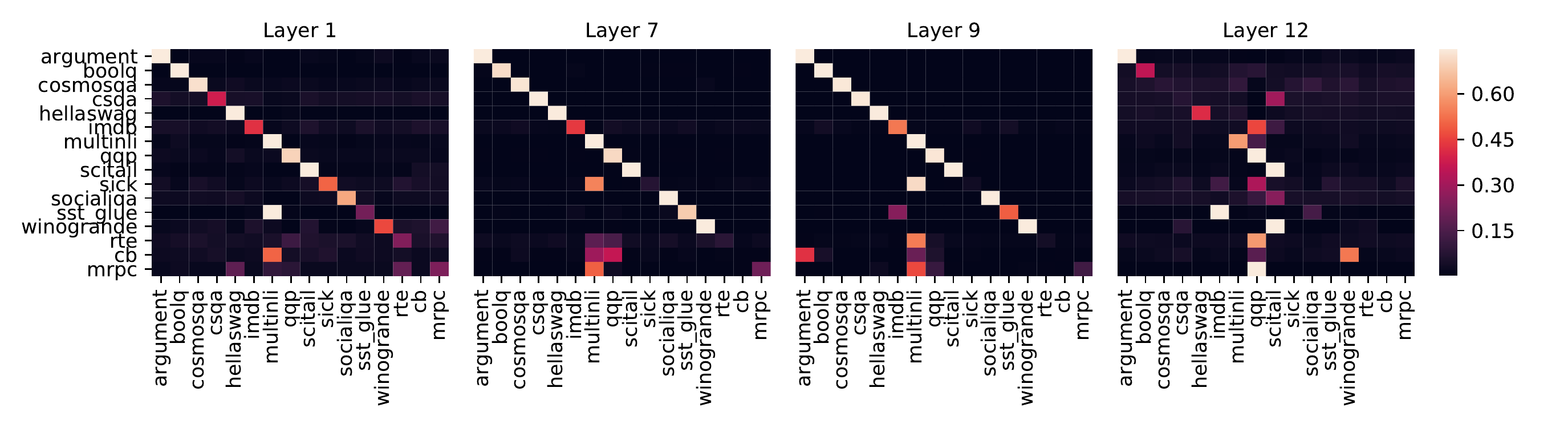}
\caption{AdapterFusion activations of pretrained \textbf{ST-Adapters}. Rows indicate the target task $m$, columns indicate adapters $n$. We
assume that the softmax activation for $\Phi_{n,l}$ is high if the information of adapter $n$ is useful for task $m$. 
For our analysis, we calculate the softmax activation for each adapter $\Phi_{n,l}$, where $n \in \{1,\ldots, N\}$, and average 
over  
all activations
within the same layer $l$ 
calculated over all instances in the development set.  }
\label{fig:Heatmap_part_mt}
\end{figure*}

\textbf{(1)} In the case of Fusion with ST-A, for $15/16$ tasks, the performance remains the same or improves as compared to the task's pretrained adapter. For $10/16$ tasks we see performance gains. This shows that having access to adapters from other tasks is beneficial and in the majority of cases leads to better results on the target task.
\textbf{(2)} We find that for $11/16$ tasks, Fusion with ST-A improves the performance compared to MT-A. This demonstrates the ability of Fusion with ST-A to share information between tasks while avoiding the interference that multi-task training suffers from. 
\textbf{(3)} For only $7/16$ tasks, we see an improvement of Fusion with MT-A over the ST-A. Training of \mbox{MT-A} in the first stage of our algorithm suffers from all the problems of multi-task learning and results in less effective adapters than our ST-A on average. Fusion helps bridge some of this gap but is not able to mitigate the entire performance drop.
\textbf{(4)} In the case of AdapterFusion with MT-A, we see that the performances on \emph{all 16 tasks} improves or stays the same. This demonstrates that AdapterFusion can successfully combine the specific adapter weights, even if the adapters were trained in a multi-task setting,  confirming that our method is  versatile.

\paragraph{Summary.}
Our findings demonstrate that Fusion with ST-A is the most promising approach to sharing information across tasks. Our approach allows us to train adapters in parallel and it requires no heuristic sampling strategies to deal with imbalanced datasets. It also allows researchers to easily add more tasks as they become available, without requiring complete model retraining. 

While Fusion with MT-A does provide gains over simply using MT-A, the effort required to train these in a multi-task setting followed by the Fusion step are not warranted by the limited gains in performance. On the other hand, we find that Fusion with ST-A is an efficient and versatile approach to transfer learning.

\section{Analysis of Fusion Activation} 
\label{analysis}

We analyze the weighting patterns that are learned by AdapterFusion to better understand which tasks impact the model predictions, and whether there exist differences across BERT layers.

We plot the  
results for layers 1, 7, 9, and 12
and ST-A in Figure \ref{fig:Heatmap_part_mt} (see Appendix Figure~\ref{fig:Heatmap_full} for the remaining layers). 
We find that   
tasks which do not benefit from AdapterFusion tend to more strongly activate their own adapter at every layer  
(e.g. \textit{Argument, HellaSwag, MNLI, QQP, SciTail}). This confirms that AdapterFusion 
only extracts information from adapters if they are beneficial for the target task $m$.
We further find that \textit{MNLI} is a useful intermediate task that benefits a large number of target tasks, e.g. \textit{BoolQ, SICK, CSQA, SST-2, CB, MRPC, RTE}, which is in line with previous work \cite{phang2018sentence, conneau2018senteval, reimers-gurevych-2019-sentence}. Similarly, \textit{QQP} is utilized by a large number of tasks, e.g. \textit{SICK, IMDB, RTE, CB, MRPC, SST-2}. Most importantly, tasks with small datasets such as \textit{CB, RTE,} and \textit{MRPC} often strongly rely on adapters trained on large datasets such as  \textit{MNLI} and \textit{QQP}. 

Interestingly, we find that the activations in layer 12 are considerably more distributed across multiple tasks than adapters in earlier layers. The potential reason for this is that the last adapters are not encapsulated between frozen pretrained layers, and can thus be considered as an extension of the prediction head. The representations of the adapters in the 12$^\text{th}$ layer  might thus not be as comparable, resulting in more distributed activations. This is in line with \citet{Pfeiffer20Gib} who are able to improve zero-shot cross-lingual performance considerably by dropping the adapters in the last layer.

\section{Contemporary Work}

In contemporaneous work, other approaches for parameter efficient fine-tuning have been proposed. 
\citet{Guo2020DiffPruning} train sparse ``diff" vectors which are applied on top of pretrained frozen parameter vectors. 
\citet{ravfogel2021bitfit} only fine-tune bias terms of the pretrained language models, achieving similar results as full model fine-tuning.
\citet{li2021prefix} propose prefix-tuning for natural language generation tasks. Here, continuous task-specific vectors are trained while  the remaining model is kept frozen. These alternative, parameter-efficient fine-tuning strategies all encapsulate the idiosyncratic task-specific information in designated parameters, creating the potential for new composition approaches of multiple tasks. 

\citet{rueckle2020adapterdrop} analyse the training and inference efficiency of adapters and AdapterFusion. For AdapterFusion, they find that adding more tasks to the set of adapters results in a linear increase of computational cost, both for training and inference. They further propose approaches to mitigate this overhead. 

\section{Conclusion and Outlook}

\subsection{Conclusion}
\label{conclusion}
We propose a novel approach to transfer learning called AdapterFusion which provides a simple and effective way to combine information from several tasks. 
By separating the extraction of knowledge from its composition,
we are able to effectively avoid the common pitfalls of multi-task learning, such as catastrophic forgetting and interference between tasks.
Further, AdapterFusion mitigates the problem of traditional multi-task learning in which complete re-training is required, when new tasks are added to the pool of datasets.

We have shown that AdapterFusion is compatible with adapters trained in both single-task as well as multi-task setups. AdapterFusion consistently outperforms fully fine-tuned models  on the target task, demonstrating the value in having access to information from other tasks. While we observe gains using both ST-A as well as MT-A, we find that composing ST-A using AdapterFusion is the more efficient strategy, as adapters can be trained in parallel and re-used.

Finally, we analyze the  weighting patterns of individual adapters in AdapterFusion which reveal that tasks with small datasets more often rely on information from tasks with large datasets, thereby achieving the largest performance gains in our experiments. We  show that AdapterFusion is able to identify and select adapters that contain knowledge relevant to task of interest, while ignoring the remaining ones. This provides an implicit no-op option and makes AdapterFusion a suitable and versatile transfer learning approach for any NLU setting.

\subsection{Outlook}

\citet{rueckle2020adapterdrop} have studied pruning a large portion of adapters after Fusion training. Their results show that removing the less activated adapters results in almost no performance drop at inference time while considerably improving the inference speed. They also provide some initial evidence that it is possible to train Fusion with a subset of the available adapters in each minibatch, potentially enabling us to scale our approach to large adapter sets~---~which would otherwise be computationally infeasible. We believe that such extensions are a promising direction for future work.

\citet{Pfeiffer20Gib} have achieved considerable improvements in the zero-shot cross-lingual transfer performance by dropping the adapters in the last layer. In preliminary results, we have observed similar trends  
with AdapterFusion when the adapters in the last layer are not used. 
We will investigate this further in future work.

\section*{Acknowledgments}
Jonas is supported by the LOEWE initiative (Hesse, Germany) within the emergenCITY center. Aishwarya  was supported in part by a DeepMind PhD Fellowship during the time which this project was carried out. Andreas  is supported by the German Research Foundation within the project “Open Argument Mining” (GU 798/25-1), associated with the Priority Program “Robust Argumentation Machines (RATIO)” (SPP-1999). This work was partly supported by Samsung Advanced Institute of Technology (Next Generation Deep Learning: from pattern recognition to AI) and Samsung Research (Improving Deep Learning using Latent Structure). Kyunghyun was a research scientist at Facebook AI Research part-time during which this project was carried out.

We thank Sebastian Ruder, Max Glockner, Jason Phang, Alex Wang, Katrina Evtimova and Sam Bowman for insightful feedback and suggestions on drafts of this paper.

\bibliography{anthology,eacl2021}

\begin{thebibliography}{64}
\expandafter\ifx\csname natexlab\endcsname\relax\def\natexlab#1{#1}\fi

\bibitem[{Ba et~al.(2016)Ba, Kiros, and Hinton}]{Ba16_LayerNorm}
Lei~Jimmy Ba, Jamie~Ryan Kiros, and Geoffrey~E. Hinton. 2016.
\newblock \href {http://arxiv.org/abs/1607.06450} {Layer normalization}.
\newblock \emph{arXiv preprint}.

\bibitem[{Bahdanau et~al.(2015)Bahdanau, Cho, and Bengio}]{Bahdanau14Attention}
Dzmitry Bahdanau, Kyunghyun Cho, and Yoshua Bengio. 2015.
\newblock \href {http://arxiv.org/abs/1409.0473} {Neural machine translation by
  jointly learning to align and translate}.
\newblock In \emph{3rd International Conference on Learning Representations,
  {ICLR} 2015, San Diego, CA, USA, May 7-9, 2015, Conference Track
  Proceedings}.

\bibitem[{Bapna and Firat(2019)}]{bapna2019simple}
Ankur Bapna and Orhan Firat. 2019.
\newblock \href {https://doi.org/10.18653/v1/D19-1165} {Simple, scalable
  adaptation for neural machine translation}.
\newblock In \emph{Proceedings of the 2019 Conference on Empirical Methods in
  Natural Language Processing and the 9th International Joint Conference on
  Natural Language Processing, {EMNLP-IJCNLP} 2019, Hong Kong, China, November
  3-7, 2019}, pages 1538--1548.

\bibitem[{Caruana(1997)}]{caruana1997multitask}
Rich Caruana. 1997.
\newblock \href {https://doi.org/10.1023/A:1007379606734} {Multitask learning}.
\newblock \emph{Machine Learning}, 28(1):41--75.

\bibitem[{Clark et~al.(2019)Clark, Lee, Chang, Kwiatkowski, Collins, and
  Toutanova}]{Clark19BoolQ}
Christopher Clark, Kenton Lee, Ming{-}Wei Chang, Tom Kwiatkowski, Michael
  Collins, and Kristina Toutanova. 2019.
\newblock \href {https://doi.org/10.18653/v1/n19-1300} {Boolq: Exploring the
  surprising difficulty of natural yes/no questions}.
\newblock In \emph{Proceedings of the 2019 Conference of the North American
  Chapter of the Association for Computational Linguistics: Human Language
  Technologies, {NAACL-HLT} 2019, Minneapolis, MN, USA, June 2-7, 2019, Volume
  1 (Long and Short Papers)}, pages 2924--2936.

\bibitem[{Collobert and Weston(2008)}]{collobert2008unified}
Ronan Collobert and Jason Weston. 2008.
\newblock \href {https://doi.org/10.1145/1390156.1390177} {A unified
  architecture for natural language processing: deep neural networks with
  multitask learning}.
\newblock In \emph{Machine Learning, Proceedings of the Twenty-Fifth
  International Conference {(ICML} 2008), Helsinki, Finland, June 5-9, 2008},
  pages 160--167.

\bibitem[{Conneau and Kiela(2018)}]{conneau2018senteval}
Alexis Conneau and Douwe Kiela. 2018.
\newblock \href
  {http://www.lrec-conf.org/proceedings/lrec2018/summaries/757.html} {Senteval:
  An evaluation toolkit for universal sentence representations}.
\newblock In \emph{Proceedings of the Eleventh International Conference on
  Language Resources and Evaluation, {LREC} 2018, Miyazaki, Japan, May 7-12,
  2018}.

\bibitem[{De~Marneffe et~al.(2019)De~Marneffe, Simons, and
  Tonhauser}]{deMarneffe2019commitmentbank}
Marie-Catherine De~Marneffe, Mandy Simons, and Judith Tonhauser. 2019.
\newblock The commitmentbank: Investigating projection in naturally occurring
  discourse.
\newblock In \emph{proceedings of Sinn und Bedeutung}, volume~23, pages
  107--124.

\bibitem[{Devlin et~al.(2019)Devlin, Chang, Lee, and
  Toutanova}]{devlin2018bert}
Jacob Devlin, Ming-Wei Chang, Kenton Lee, and Kristina Toutanova. 2019.
\newblock \href {https://doi.org/10.18653/v1/N19-1423} {{BERT}: Pre-training of
  deep bidirectional transformers for language understanding}.
\newblock In \emph{Proceedings of the 2019 Conference of the North {A}merican
  Chapter of the Association for Computational Linguistics: Human Language
  Technologies, Volume 1 (Long and Short Papers)}, pages 4171--4186,
  Minneapolis, Minnesota. Association for Computational Linguistics.

\bibitem[{Dolan and Brockett(2005)}]{Dolan05MRPC}
William~B. Dolan and Chris Brockett. 2005.
\newblock \href {https://www.aclweb.org/anthology/I05-5002/} {Automatically
  constructing a corpus of sentential paraphrases}.
\newblock In \emph{Proceedings of the Third International Workshop on
  Paraphrasing, IWP@IJCNLP 2005, Jeju Island, Korea, October 2005, 2005}.

\bibitem[{French(1999)}]{french1999catastrophic}
Robert~M French. 1999.
\newblock Catastrophic forgetting in connectionist networks.
\newblock \emph{Trends in cognitive sciences}, 3(4):128--135.

\bibitem[{Guo et~al.(2020)Guo, Rush, and Kim}]{Guo2020DiffPruning}
Demi Guo, Alexander~M. Rush, and Yoon Kim. 2020.
\newblock \href {http://arxiv.org/abs/2012.07463} {Parameter-efficient transfer
  learning with diff pruning}.
\newblock \emph{arXiv preprint}.

\bibitem[{Houlsby et~al.(2019)Houlsby, Giurgiu, Jastrz{k{e}}bski, Morrone,
  de~Laroussilhe, Gesmundo, Attariyan, and Gelly}]{houlsby2019parameter}
Neil Houlsby, Andrei Giurgiu, Stanis{l}aw Jastrz{k{e}}bski, Bruna Morrone,
  Quentin de~Laroussilhe, Andrea Gesmundo, Mona Attariyan, and Sylvain Gelly.
  2019.
\newblock \href {http://proceedings.mlr.press/v97/houlsby19a.html}
  {Parameter-efficient transfer learning for {NLP}}.
\newblock In \emph{Proceedings of the 36th International Conference on Machine
  Learning, {ICML} 2019, 9-15 June 2019, Long Beach, California, {USA}}, pages
  2790--2799.

\bibitem[{Howard and Ruder(2018)}]{howard2018universal}
Jeremy Howard and Sebastian Ruder. 2018.
\newblock \href {https://doi.org/10.18653/v1/P18-1031} {Universal language
  model fine-tuning for text classification}.
\newblock In \emph{Proceedings of the 56th Annual Meeting of the Association
  for Computational Linguistics, {ACL} 2018, Melbourne, Australia, July 15-20,
  2018, Volume 1: Long Papers}, pages 328--339.

\bibitem[{Huang et~al.(2019)Huang, Bras, Bhagavatula, and
  Choi}]{Huang19cosmosqa}
Lifu Huang, Ronan~Le Bras, Chandra Bhagavatula, and Yejin Choi. 2019.
\newblock \href {https://doi.org/10.18653/v1/D19-1243} {Cosmos {QA:} machine
  reading comprehension with contextual commonsense reasoning}.
\newblock In \emph{Proceedings of the 2019 Conference on Empirical Methods in
  Natural Language Processing and the 9th International Joint Conference on
  Natural Language Processing, {EMNLP-IJCNLP} 2019, Hong Kong, China, November
  3-7, 2019}, pages 2391--2401.

\bibitem[{Khot et~al.(2018)Khot, Sabharwal, and Clark}]{Khot18SciTail}
Tushar Khot, Ashish Sabharwal, and Peter Clark. 2018.
\newblock \href
  {https://www.aaai.org/ocs/index.php/AAAI/AAAI18/paper/view/17368} {Scitail:
  {A} textual entailment dataset from science question answering}.
\newblock In \emph{Proceedings of the Thirty-Second {AAAI} Conference on
  Artificial Intelligence, (AAAI-18), the 30th innovative Applications of
  Artificial Intelligence (IAAI-18), and the 8th {AAAI} Symposium on
  Educational Advances in Artificial Intelligence (EAAI-18), New Orleans,
  Louisiana, USA, February 2-7, 2018}, pages 5189--5197.

\bibitem[{Lauscher et~al.(2020)Lauscher, Majewska, Ribeiro, Gurevych, Rozanov,
  and Glava\v{s}}]{Lauscher2020comonsense}
Anne Lauscher, Olga Majewska, Leonardo F.~R. Ribeiro, Iryna Gurevych, Nikolai
  Rozanov, and Goran Glava\v{s}. 2020.
\newblock \href {https://arxiv.org/abs/2005.11787} {{Common Sense or World
  Knowledge? Investigating Adapter-Based Knowledge Injection into Pretrained
  Transformers}}.
\newblock \emph{arXiv preprint}.

\bibitem[{Lee et~al.(2017)Lee, Cho, and Hofmann}]{lee2017fully}
Jason Lee, Kyunghyun Cho, and Thomas Hofmann. 2017.
\newblock \href {https://transacl.org/ojs/index.php/tacl/article/view/1051}
  {Fully character-level neural machine translation without explicit
  segmentation}.
\newblock \emph{Transactions of the Association for Computational Linguistics
  2017}, 5:365--378.

\bibitem[{Levesque(2011)}]{Levesque11winograd}
Hector~J. Levesque. 2011.
\newblock \href {http://www.aaai.org/ocs/index.php/SSS/SSS11/paper/view/2502}
  {The winograd schema challenge}.
\newblock In \emph{Logical Formalizations of Commonsense Reasoning, Papers from
  the 2011 {AAAI} Spring Symposium, Technical Report SS-11-06, Stanford,
  California, USA, March 21-23, 2011}.

\bibitem[{Li and Liang(2021)}]{li2021prefix}
Xiang~Lisa Li and Percy Liang. 2021.
\newblock \href {https://arxiv.org/pdf/2101.00190.pdf} {Prefix-tuning:
  Optimizing continuous prompts for generation}.
\newblock \emph{arXiv preprint}.

\bibitem[{Liu et~al.(2016)Liu, Qiu, and Huang}]{liu2016recurrent}
Pengfei Liu, Xipeng Qiu, and Xuanjing Huang. 2016.
\newblock \href {http://www.ijcai.org/Abstract/16/408} {Recurrent neural
  network for text classification with multi-task learning}.
\newblock In \emph{Proceedings of the Twenty-Fifth International Joint
  Conference on Artificial Intelligence, {IJCAI} 2016, New York, NY, USA, 9-15
  July 2016}, pages 2873--2879.

\bibitem[{Liu et~al.(2017)Liu, Qiu, and Huang}]{liu2017adversarial}
Pengfei Liu, Xipeng Qiu, and Xuanjing Huang. 2017.
\newblock \href {https://doi.org/10.18653/v1/P17-1001} {Adversarial multi-task
  learning for text classification}.
\newblock In \emph{Proceedings of the 55th Annual Meeting of the Association
  for Computational Linguistics (Volume 1: Long Papers)}, pages 1--10,
  Vancouver, Canada. Association for Computational Linguistics.

\bibitem[{Liu et~al.(2019{\natexlab{a}})Liu, He, Chen, and Gao}]{liu2019multi}
Xiaodong Liu, Pengcheng He, Weizhu Chen, and Jianfeng Gao. 2019{\natexlab{a}}.
\newblock \href {https://doi.org/10.18653/v1/p19-1441} {Multi-task deep neural
  networks for natural language understanding}.
\newblock In \emph{Proceedings of the 57th Conference of the Association for
  Computational Linguistics, {ACL} 2019, Florence, Italy, July 28- August 2,
  2019, Volume 1: Long Papers}, pages 4487--4496.

\bibitem[{Liu et~al.(2019{\natexlab{b}})Liu, Ott, Goyal, Du, Joshi, Chen, Levy,
  Lewis, Zettlemoyer, and Stoyanov}]{liu2019roberta}
Yinhan Liu, Myle Ott, Naman Goyal, Jingfei Du, Mandar Joshi, Danqi Chen, Omer
  Levy, Mike Lewis, Luke Zettlemoyer, and Veselin Stoyanov. 2019{\natexlab{b}}.
\newblock \href {https://arxiv.org/abs/1907.11692} {Roberta: A robustly
  optimized bert pretraining approach}.
\newblock \emph{arXiv preprint}.

\bibitem[{Maas et~al.(2011)Maas, Daly, Pham, Huang, Ng, and Potts}]{Maas11IMDB}
Andrew~L. Maas, Raymond~E. Daly, Peter~T. Pham, Dan Huang, Andrew~Y. Ng, and
  Christopher Potts. 2011.
\newblock \href {https://www.aclweb.org/anthology/P11-1015/} {Learning word
  vectors for sentiment analysis}.
\newblock In \emph{The 49th Annual Meeting of the Association for Computational
  Linguistics: Human Language Technologies, Proceedings of the Conference,
  19-24 June, 2011, Portland, Oregon, {USA}}, pages 142--150.

\bibitem[{Marelli et~al.(2014)Marelli, Menini, Baroni, Bentivogli, Bernardi,
  and Zamparelli}]{marelli-etal-2014-sick}
Marco Marelli, Stefano Menini, Marco Baroni, Luisa Bentivogli, Raffaella
  Bernardi, and Roberto Zamparelli. 2014.
\newblock \href
  {http://www.lrec-conf.org/proceedings/lrec2014/pdf/363_Paper.pdf} {A {SICK}
  cure for the evaluation of compositional distributional semantic models}.
\newblock In \emph{Proceedings of the Ninth International Conference on
  Language Resources and Evaluation ({LREC}-2014)}, pages 216--223, Reykjavik,
  Iceland. European Languages Resources Association (ELRA).

\bibitem[{McCloskey and Cohen(1989)}]{mccloskey1989catastrophic}
Michael McCloskey and Neal~J Cohen. 1989.
\newblock Catastrophic interference in connectionist networks: The sequential
  learning problem.
\newblock In \emph{Psychology of learning and motivation}, volume~24, pages
  109--165. Elsevier.

\bibitem[{Nam et~al.(2014)Nam, Kim, Loza~Menc'{i}a, Gurevych, and
  F{\"{u}}rnkranz}]{nam2014large}
Jinseok Nam, Jungi Kim, Eneldo Loza~Menc'{i}a, Iryna Gurevych, and Johannes
  F{\"{u}}rnkranz. 2014.
\newblock \href {https://doi.org/10.1007/978-3-662-44851-9\_28} {Large-scale
  multi-label text classification - revisiting neural networks}.
\newblock In \emph{Machine Learning and Knowledge Discovery in Databases -
  European Conference, {ECML} {PKDD} 2014, Nancy, France, September 15-19,
  2014. Proceedings, Part {II}}, pages 437--452.

\bibitem[{Pan and Yang(2010)}]{pan2009survey}
Sinno~Jialin Pan and Qiang Yang. 2010.
\newblock \href {https://doi.org/10.1109/TKDE.2009.191} {A survey on transfer
  learning}.
\newblock \emph{{IEEE} Trans. Knowl. Data Eng.}, 22(10):1345--1359.

\bibitem[{Peters et~al.(2019)Peters, Ruder, and Smith}]{peters2019tune}
Matthew~E. Peters, Sebastian Ruder, and Noah~A. Smith. 2019.
\newblock \href {https://doi.org/10.18653/v1/w19-4302} {To tune or not to tune?
  adapting pretrained representations to diverse tasks}.
\newblock In \emph{Proceedings of the 4th Workshop on Representation Learning
  for NLP, RepL4NLP@ACL 2019, Florence, Italy, August 2, 2019}, pages 7--14.

\bibitem[{Pfeiffer et~al.(2020{\natexlab{a}})Pfeiffer, R{\"u}ckl{\'e}, Poth,
  Kamath, Vuli{\'c}, Ruder, Cho, and Gurevych}]{pfeiffer2020AdapterHub}
Jonas Pfeiffer, Andreas R{\"u}ckl{\'e}, Clifton Poth, Aishwarya Kamath, Ivan
  Vuli{\'c}, Sebastian Ruder, Kyunghyun Cho, and Iryna Gurevych.
  2020{\natexlab{a}}.
\newblock \href {https://www.aclweb.org/anthology/2020.emnlp-demos.7}
  {{A}dapter{H}ub: A framework for adapting transformers}.
\newblock In \emph{Proceedings of the 2020 Conference on Empirical Methods in
  Natural Language Processing: System Demonstrations}, pages 46--54, Online.
  Association for Computational Linguistics.

\bibitem[{Pfeiffer et~al.(2020{\natexlab{b}})Pfeiffer, Simpson, and
  Gurevych}]{pfeiffer20multilabel}
Jonas Pfeiffer, Edwin Simpson, and Iryna Gurevych. 2020{\natexlab{b}}.
\newblock \href {https://arxiv.org/pdf/2005.00250.pdf} {Low resource multi-task
  sequence tagging - revisiting dynamic conditional random fields}.
\newblock \emph{arXiv preprint}.

\bibitem[{Pfeiffer et~al.(2020{\natexlab{c}})Pfeiffer, Vuli{\'c}, Gurevych, and
  Ruder}]{pfeiffer20madx}
Jonas Pfeiffer, Ivan Vuli{\'c}, Iryna Gurevych, and Sebastian Ruder.
  2020{\natexlab{c}}.
\newblock \href {https://www.aclweb.org/anthology/2020.emnlp-main.617}
  {{MAD-X}: {A}n {A}dapter-{B}ased {F}ramework for {M}ulti-{T}ask
  {C}ross-{L}ingual {T}ransfer}.
\newblock In \emph{Proceedings of the 2020 Conference on Empirical Methods in
  Natural Language Processing (EMNLP)}, pages 7654--7673, Online. Association
  for Computational Linguistics.

\bibitem[{Pfeiffer et~al.(2020{\natexlab{d}})Pfeiffer, Vuli\'{c}, Gurevych, and
  Ruder}]{Pfeiffer20Gib}
Jonas Pfeiffer, Ivan Vuli\'{c}, Iryna Gurevych, and Sebastian Ruder.
  2020{\natexlab{d}}.
\newblock \href {https://arxiv.org/pdf/2012.15562.pdf} {{UNKs Everywhere:
  Adapting Multilingual Language Models to New Scripts}}.
\newblock \emph{arXiv preprint}.

\bibitem[{Phang et~al.(2018)Phang, F{\'{e}}vry, and Bowman}]{phang2018sentence}
Jason Phang, Thibault F{\'{e}}vry, and Samuel~R. Bowman. 2018.
\newblock \href {http://arxiv.org/abs/1811.01088} {Sentence encoders on stilts:
  Supplementary training on intermediate labeled-data tasks}.
\newblock \emph{arXiv preprint}.

\bibitem[{Philip et~al.(2020)Philip, Berard, Gall{\'{e}}, and
  Besacier}]{Philip20NaverAdapter}
Jerin Philip, Alexandre Berard, Matthias Gall{\'{e}}, and Laurent Besacier.
  2020.
\newblock \href {https://doi.org/10.18653/v1/2020.emnlp-main.361} {Monolingual
  adapters for zero-shot neural machine translation}.
\newblock In \emph{Proceedings of the 2020 Conference on Empirical Methods in
  Natural Language Processing, {EMNLP} 2020, Online, November 16-20, 2020},
  pages 4465--4470.

\bibitem[{Pruksachatkun et~al.(2020)Pruksachatkun, Phang, Liu, Htut, Zhang,
  Pang, Vania, Kann, and Bowman}]{pruksachatkun2020intermediate}
Yada Pruksachatkun, Jason Phang, Haokun Liu, Phu~Mon Htut, Xiaoyi Zhang,
  Richard~Yuanzhe Pang, Clara Vania, Katharina Kann, and Samuel Bowman. 2020.
\newblock \href {https://arxiv.org/abs/2005.00628} {Intermediate-task transfer
  learning with pretrained language models: When and why does it work?}
\newblock In \emph{Proceedings of the 58th Annual Meeting of the Association
  for Computational Linguistics}, pages 5231--5247.

\bibitem[{Radford et~al.(2018)Radford, Narasimhan, Salimans, and
  Sutskever}]{radford2018improving}
Alec Radford, Karthik Narasimhan, Tim Salimans, and Ilya Sutskever. 2018.
\newblock \href
  {https://www.cs.ubc.ca/~amuham01/LING530/papers/radford2018improving.pdf}
  {Improving language understanding by generative pre-training}.

\bibitem[{Ravfogel and Goldberg(2021)}]{ravfogel2021bitfit}
Elad Ben-Zaken1~Shauli Ravfogel and Yoav Goldberg. 2021.
\newblock \href {https://nlp.biu.ac.il/~yogo/bitfit.pdf} {Bitfit: Simple
  parameter-efficient fine-tuning for transformer-based masked
  language-models}.
\newblock \emph{arXiv preprint}.

\bibitem[{Rebuffi et~al.(2017)Rebuffi, Bilen, and
  Vedaldi}]{rebuffi2017learning}
Sylvestre{-}Alvise Rebuffi, Hakan Bilen, and Andrea Vedaldi. 2017.
\newblock \href
  {http://papers.nips.cc/paper/6654-learning-multiple-visual-domains-with-residual-adapters}
  {Learning multiple visual domains with residual adapters}.
\newblock In \emph{Advances in Neural Information Processing Systems 30: Annual
  Conference on Neural Information Processing Systems 2017, 4-9 December 2017,
  Long Beach, CA, {USA}}, pages 506--516.

\bibitem[{Reimers and Gurevych(2019)}]{reimers-gurevych-2019-sentence}
Nils Reimers and Iryna Gurevych. 2019.
\newblock \href {https://doi.org/10.18653/v1/D19-1410} {Sentence-{BERT}:
  Sentence embeddings using {S}iamese {BERT}-networks}.
\newblock In \emph{Proceedings of the 2019 Conference on Empirical Methods in
  Natural Language Processing and the 9th International Joint Conference on
  Natural Language Processing (EMNLP-IJCNLP)}, pages 3980--3990, Hong Kong,
  China. Association for Computational Linguistics.

\bibitem[{R{\"u}ckl{\'e} et~al.(2020{\natexlab{a}})R{\"u}ckl{\'e}, Geigle,
  Glockner, Beck, Pfeiffer, Reimers, and Gurevych}]{rueckle2020adapterdrop}
Andreas R{\"u}ckl{\'e}, Gregor Geigle, Max Glockner, Tilman Beck, Jonas
  Pfeiffer, Nils Reimers, and Iryna Gurevych. 2020{\natexlab{a}}.
\newblock \href {https://arxiv.org/pdf/2010.11918.pdf} {{AdapterDrop: On the
  Efficiency of Adapters in Transformers}}.
\newblock \emph{arXiv preprint}.

\bibitem[{R{\"u}ckl{\'e} et~al.(2020{\natexlab{b}})R{\"u}ckl{\'e}, Pfeiffer,
  and Gurevych}]{rueckle2020MultiCQA}
Andreas R{\"u}ckl{\'e}, Jonas Pfeiffer, and Iryna Gurevych. 2020{\natexlab{b}}.
\newblock \href {https://doi.org/10.18653/v1/2020.emnlp-main.194}
  {{M}ulti{CQA}: Zero-shot transfer of self-supervised text matching models on
  a massive scale}.
\newblock In \emph{Proceedings of the 2020 Conference on Empirical Methods in
  Natural Language Processing (EMNLP)}, pages 2471--2486, Online. Association
  for Computational Linguistics.

\bibitem[{Ruder(2017)}]{ruder2017}
Sebastian Ruder. 2017.
\newblock \href {http://arxiv.org/abs/1706.05098} {An overview of multi-task
  learning in deep neural networks}.
\newblock \emph{arXiv preprint}.

\bibitem[{Ruder(2019)}]{ruder2019neural}
Sebastian Ruder. 2019.
\newblock \emph{Neural Transfer Learning for Natural Language Processing}.
\newblock Ph.D. thesis, National University of Ireland, Galway.

\bibitem[{Ruder et~al.(2019)Ruder, Bingel, Augenstein, and
  S{\o}gaard}]{ruder2019latent}
Sebastian Ruder, Joachim Bingel, Isabelle Augenstein, and Anders S{\o}gaard.
  2019.
\newblock \href {https://doi.org/10.1609/aaai.v33i01.33014822} {Latent
  multi-task architecture learning}.
\newblock In \emph{The Thirty-Third {AAAI} Conference on Artificial
  Intelligence, {AAAI} 2019, The Thirty-First Innovative Applications of
  Artificial Intelligence Conference, {IAAI} 2019, The Ninth {AAAI} Symposium
  on Educational Advances in Artificial Intelligence, {EAAI} 2019, Honolulu,
  Hawaii, USA, January 27 - February 1, 2019}, pages 4822--4829.

\bibitem[{Sakaguchi et~al.(2020)Sakaguchi, Bras, Bhagavatula, and
  Choi}]{Sakaguchi19winogrande}
Keisuke Sakaguchi, Ronan~Le Bras, Chandra Bhagavatula, and Yejin Choi. 2020.
\newblock \href {https://aaai.org/ojs/index.php/AAAI/article/view/6399}
  {Winogrande: An adversarial winograd schema challenge at scale}.
\newblock In \emph{The Thirty-Fourth {AAAI} Conference on Artificial
  Intelligence, {AAAI} 2020, The Thirty-Second Innovative Applications of
  Artificial Intelligence Conference, {IAAI} 2020, The Tenth {AAAI} Symposium
  on Educational Advances in Artificial Intelligence, {EAAI} 2020, New York,
  NY, USA, February 7-12, 2020}, pages 8732--8740.

\bibitem[{Sanh et~al.(2019)Sanh, Wolf, and Ruder}]{sanh2019hierarchical}
Victor Sanh, Thomas Wolf, and Sebastian Ruder. 2019.
\newblock \href {https://doi.org/10.1609/aaai.v33i01.33016949} {A hierarchical
  multi-task approach for learning embeddings from semantic tasks}.
\newblock In \emph{The Thirty-Third {AAAI} Conference on Artificial
  Intelligence, {AAAI} 2019, The Thirty-First Innovative Applications of
  Artificial Intelligence Conference, {IAAI} 2019, The Ninth {AAAI} Symposium
  on Educational Advances in Artificial Intelligence, {EAAI} 2019, Honolulu,
  Hawaii, USA, January 27 - February 1, 2019}, pages 6949--6956.

\bibitem[{Sap et~al.(2019)Sap, Rashkin, Chen, Bras, and Choi}]{Sap19SocialIQA}
Maarten Sap, Hannah Rashkin, Derek Chen, Ronan~Le Bras, and Yejin Choi. 2019.
\newblock \href {https://doi.org/10.18653/v1/D19-1454} {Social iqa: Commonsense
  reasoning about social interactions}.
\newblock In \emph{Proceedings of the 2019 Conference on Empirical Methods in
  Natural Language Processing and the 9th International Joint Conference on
  Natural Language Processing, {EMNLP-IJCNLP} 2019, Hong Kong, China, November
  3-7, 2019}, pages 4462--4472.

\bibitem[{Socher et~al.(2013)Socher, Perelygin, Wu, Chuang, Manning, Ng, and
  Potts}]{socher-etal-2013-recursive}
Richard Socher, Alex Perelygin, Jean Wu, Jason Chuang, Christopher~D. Manning,
  Andrew Ng, and Christopher Potts. 2013.
\newblock \href {https://www.aclweb.org/anthology/D13-1170} {Recursive deep
  models for semantic compositionality over a sentiment treebank}.
\newblock In \emph{Proceedings of the 2013 Conference on Empirical Methods in
  Natural Language Processing}, pages 1631--1642, Seattle, Washington, USA.
  Association for Computational Linguistics.

\bibitem[{Speer et~al.(2017)Speer, Chin, and Havasi}]{SpeerCH17ConceptNet}
Robyn Speer, Joshua Chin, and Catherine Havasi. 2017.
\newblock \href {http://aaai.org/ocs/index.php/AAAI/AAAI17/paper/view/14972}
  {Conceptnet 5.5: An open multilingual graph of general knowledge}.
\newblock In \emph{Proceedings of the Thirty-First {AAAI} Conference on
  Artificial Intelligence, February 4-9, 2017, San Francisco, California,
  {USA}}, pages 4444--4451.

\bibitem[{Stab et~al.(2018)Stab, Miller, Schiller, Rai, and
  Gurevych}]{Stab18Argument}
Christian Stab, Tristan Miller, Benjamin Schiller, Pranav Rai, and Iryna
  Gurevych. 2018.
\newblock \href {https://doi.org/10.18653/v1/d18-1402} {Cross-topic argument
  mining from heterogeneous sources}.
\newblock In \emph{Proceedings of the 2018 Conference on Empirical Methods in
  Natural Language Processing, Brussels, Belgium, October 31 - November 4,
  2018}, pages 3664--3674.

\bibitem[{Stickland and Murray(2019)}]{stickland2019bert}
Asa~Cooper Stickland and Iain Murray. 2019.
\newblock \href {http://proceedings.mlr.press/v97/stickland19a.html} {{BERT}
  and pals: Projected attention layers for efficient adaptation in multi-task
  learning}.
\newblock In \emph{Proceedings of the 36th International Conference on Machine
  Learning, {ICML} 2019, 9-15 June 2019, Long Beach, California, {USA}}, pages
  5986--5995.

\bibitem[{Talmor et~al.(2019)Talmor, Herzig, Lourie, and Berant}]{Talmor19CSQA}
Alon Talmor, Jonathan Herzig, Nicholas Lourie, and Jonathan Berant. 2019.
\newblock \href {https://doi.org/10.18653/v1/n19-1421} {Commonsenseqa: {A}
  question answering challenge targeting commonsense knowledge}.
\newblock In \emph{Proceedings of the 2019 Conference of the North American
  Chapter of the Association for Computational Linguistics: Human Language
  Technologies, {NAACL-HLT} 2019, Minneapolis, MN, USA, June 2-7, 2019, Volume
  1 (Long and Short Papers)}, pages 4149--4158.

\bibitem[{Torrey and Shavlik(2010)}]{torrey2010transfer}
Lisa Torrey and Jude Shavlik. 2010.
\newblock Transfer learning.
\newblock In \emph{Handbook of research on machine learning applications and
  trends: algorithms, methods, and techniques}, pages 242--264. IGI Global.

\bibitem[{{\"U}st{\"u}n et~al.(2020){\"U}st{\"u}n, Bisazza, Bouma, and van
  Noord}]{ustun-etal-2020-udapter}
Ahmet {\"U}st{\"u}n, Arianna Bisazza, Gosse Bouma, and Gertjan van Noord. 2020.
\newblock \href {https://www.aclweb.org/anthology/2020.emnlp-main.180}
  {{UD}apter: Language adaptation for truly {U}niversal {D}ependency parsing}.
\newblock In \emph{Proceedings of the 2020 Conference on Empirical Methods in
  Natural Language Processing (EMNLP)}, pages 2302--2315, Online. Association
  for Computational Linguistics.

\bibitem[{Vaswani et~al.(2017)Vaswani, Shazeer, Parmar, Uszkoreit, Jones,
  Gomez, Kaiser, and Polosukhin}]{Vaswani17}
Ashish Vaswani, Noam Shazeer, Niki Parmar, Jakob Uszkoreit, Llion Jones,
  Aidan~N. Gomez, Lukasz Kaiser, and Illia Polosukhin. 2017.
\newblock \href {http://papers.nips.cc/paper/7181-attention-is-all-you-need}
  {Attention is all you need}.
\newblock In \emph{Advances in Neural Information Processing Systems 30: Annual
  Conference on Neural Information Processing Systems 2017, 4-9 December 2017,
  Long Beach, CA, {USA}}, pages 5998--6008.

\bibitem[{Vidoni et~al.(2020)Vidoni, Vuli{\'c}, and
  Glava\v{s}}]{Vidoni2020OrthogonalLA}
M.~Vidoni, Ivan Vuli{\'c}, and Goran Glava\v{s}. 2020.
\newblock \href {https://arxiv.org/pdf/2012.06460.pdf} {Orthogonal language and
  task adapters in zero-shot cross-lingual transfer}.
\newblock In \emph{arXiv preprint}.

\bibitem[{Wang et~al.(2018)Wang, Singh, Michael, Hill, Levy, and
  Bowman}]{wang18glue}
Alex Wang, Amanpreet Singh, Julian Michael, Felix Hill, Omer Levy, and
  Samuel~R. Bowman. 2018.
\newblock \href {https://doi.org/10.18653/v1/w18-5446} {{GLUE:} {A} multi-task
  benchmark and analysis platform for natural language understanding}.
\newblock In \emph{Proceedings of the Workshop: Analyzing and Interpreting
  Neural Networks for NLP, BlackboxNLP@EMNLP 2018, Brussels, Belgium, November
  1, 2018}, pages 353--355.

\bibitem[{Wang et~al.(2020)Wang, Tang, Duan, Wei, Huang, Ji, Cao, Jiang, and
  Zhou}]{Wang20kadapters}
Ruize Wang, Duyu Tang, Nan Duan, Zhongyu Wei, Xuanjing Huang, Jianshu Ji,
  Guihong Cao, Daxin Jiang, and Ming Zhou. 2020.
\newblock \href {http://arxiv.org/abs/2002.01808} {K-adapter: Infusing
  knowledge into pre-trained models with adapters}.
\newblock \emph{arXiv preprint}.

\bibitem[{Williams et~al.(2018)Williams, Nangia, and Bowman}]{Williams18MNLI}
Adina Williams, Nikita Nangia, and Samuel~R. Bowman. 2018.
\newblock \href {https://doi.org/10.18653/v1/n18-1101} {A broad-coverage
  challenge corpus for sentence understanding through inference}.
\newblock In \emph{Proceedings of the 2018 Conference of the North American
  Chapter of the Association for Computational Linguistics: Human Language
  Technologies, {NAACL-HLT} 2018, New Orleans, Louisiana, USA, June 1-6, 2018,
  Volume 1 (Long Papers)}, pages 1112--1122.

\bibitem[{Zellers et~al.(2018)Zellers, Bisk, Schwartz, and
  Choi}]{zellers18swag}
Rowan Zellers, Yonatan Bisk, Roy Schwartz, and Yejin Choi. 2018.
\newblock \href {https://doi.org/10.18653/v1/d18-1009} {{SWAG:} {A} large-scale
  adversarial dataset for grounded commonsense inference}.
\newblock In \emph{Proceedings of the 2018 Conference on Empirical Methods in
  Natural Language Processing, Brussels, Belgium, October 31 - November 4,
  2018}, pages 93--104.

\bibitem[{Zellers et~al.(2019)Zellers, Holtzman, Bisk, Farhadi, and
  Choi}]{zellers19hellaswag}
Rowan Zellers, Ari Holtzman, Yonatan Bisk, Ali Farhadi, and Yejin Choi. 2019.
\newblock \href {https://doi.org/10.18653/v1/p19-1472} {Hellaswag: Can a
  machine really finish your sentence?}
\newblock In \emph{Proceedings of the 57th Conference of the Association for
  Computational Linguistics, {ACL} 2019, Florence, Italy, July 28- August 2,
  2019, Volume 1: Long Papers}, pages 4791--4800.

\bibitem[{Zhang and Yang(2017)}]{zhang2017survey}
Yu~Zhang and Qiang Yang. 2017.
\newblock \href {http://arxiv.org/abs/1707.08114} {A survey on multi-task
  learning}.
\newblock \emph{arXiv preprint}.

\end{thebibliography}
\bibliographystyle{acl_natbib}

\appendix

\section{Appendices}

\subsection{Datasets}
\label{sec:datasets_app} 
\paragraph{Commonsense Reasoning} We work with a large number of datasets, all of which have emerged recently in this domain, ranging from sentence level and document level classification to multiple choice questions. The next sentence prediction task 
\textit{HellaSWAG} \cite{zellers19hellaswag} is a more difficult version of the previously released \emph{SWAG} dataset \cite{zellers18swag}. 
\textit{Winogrande} \cite{Sakaguchi19winogrande} is a large scale and adversarially filtered \cite{zellers18swag} adaptation of the \textit{Winograd Schema Challenge} \cite{Levesque11winograd}.
\textit{Cosmos QA} \cite{Huang19cosmosqa} is a commonsense reading comprehension dataset which requires reasoning over larger text passages.
\textit{Social IQA} \cite{Sap19SocialIQA} is a multiple choice dataset which requires reasoning over social interactions between humans.
\textit{Commonsense QA} \cite{Talmor19CSQA} is a multiple choice dataset based on ConceptNet \cite{SpeerCH17ConceptNet}, which requires reasoning over general knowledge. 

\paragraph{Sentiment Analysis} We conduct experiments on two binary sentiment classification tasks on long and short text passages.
\textit{IMDb} \cite{Maas11IMDB} consists of long movie reviews 
and  
\textit{SST-2} \cite{socher-etal-2013-recursive} consists of short movie reviews from Rotten Tomatoes\footnote{\href{http://www.rottentomatoes.com/}{www.rottentomatoes.com}}.

\paragraph{Natural Language Inference (NLI)} 
The goal
 is to classify whether two sentences entail, contradict, or are neutral to each other. For this we conduct experiments on 
\textit{MultiNLI} \cite{Williams18MNLI}, a multi-genre dataset, 
\textit{SciTail} \cite{Khot18SciTail} a NLI dataset on scientific text, 
\textit{SICK} \cite{marelli-etal-2014-sick} a NLI dataset with relatedness scores,  the composition of \textit{Recognizing Textual Entailment (RTE)} datasets provided by \citet*{wang18glue}, as well as the \textit{Commitment Bank (CB)} \cite{deMarneffe2019commitmentbank} three-class textual entailment dataset. 

\paragraph{Sentence Relatedness} We include two semantic relatedness datasets which capture whether or not two text samples include similar content.
\textit{Microsoft Research Paraphrase Corpus (MRPC)} \cite{Dolan05MRPC} consists of sentence pairs which capture a paraphrase/semantic equivalence relationship. \textit{Quora Question Pairs (QQP)}  targets duplicate question detection.\footnote{\href{http://data.quora.com/First-Quora-DatasetReleaseQuestion-Pairs}{data.quora.com/First-Quora-DatasetReleaseQuestion-Pairs}}

\paragraph{Misc} 
The Argument Aspect corpus \cite{Stab18Argument} is a three-way classification task to predict whether a document provides arguments \textit{for}, \textit{against} or \textit{none} for a given topic (Nuclear Energy, Abortion, Gun-Control, etc). 
BoolQ \cite{Clark19BoolQ} is a binary reading comprehension classification task for simple \textit{yes, no} questions.

\subsection{What Is The Best Adapter Setup?} 
\label{sec:adapter_design}

 \begin{figure} 
\centering
\includegraphics[width=0.9\linewidth]{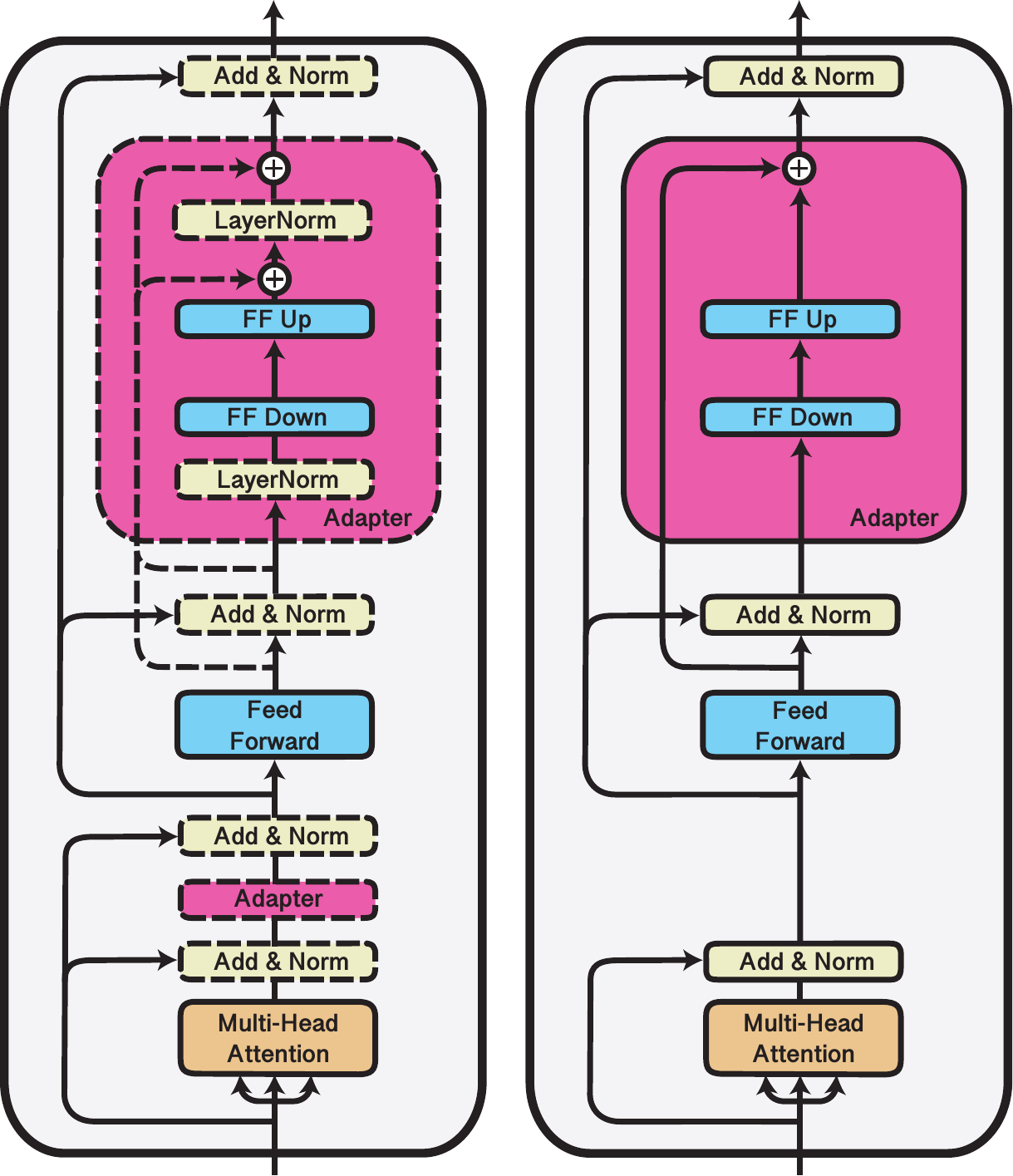}
\caption{Different architectural components of the adapter. On the left, we show 
all components for which we conduct an exhaustive search (dashed lines). On the right, we show the adapter architecture that performs the best across all our tasks. }
\label{fig:Adapter_Architecture}
\end{figure}

\begin{figure*}[htp]
\subfigure[Adapter Positions in Layer]
{\includegraphics[width=0.32\textwidth, valign=t ]{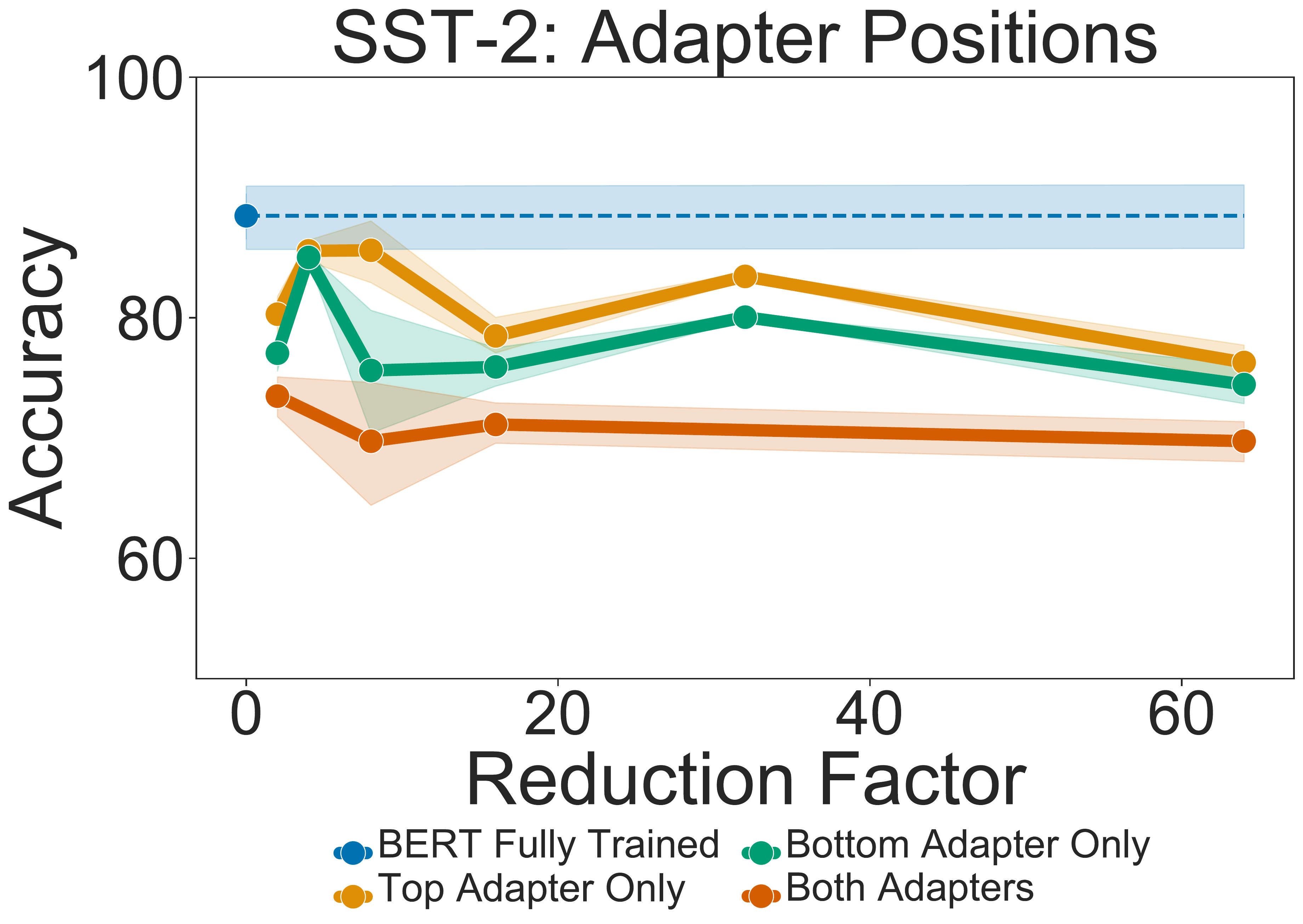}
\vphantom{\includegraphics[width=0.33\textwidth,valign=t]{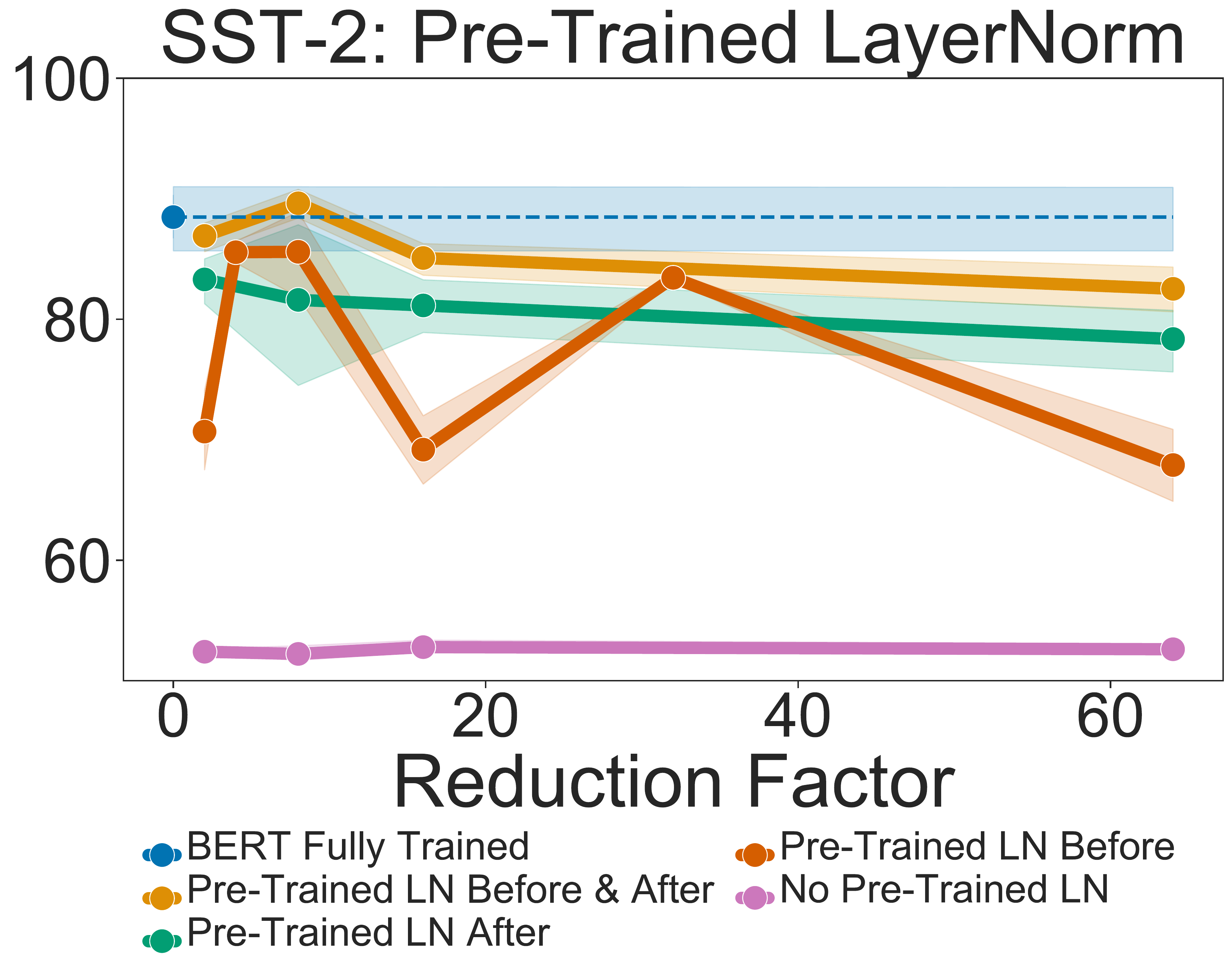}}
} \quad
\subfigure[Position of pretrained LayerNorm ]{\includegraphics[width=0.3025\textwidth, valign=t ]{img/SST_Adapters_2.pdf}
\vphantom{\includegraphics[width=0.33\textwidth,valign=t]{img/SST_Adapters_2.pdf}}}
\quad
\subfigure[Position of newly trained LayerNorm   ]{\includegraphics[width=0.3025\textwidth, valign=t ]{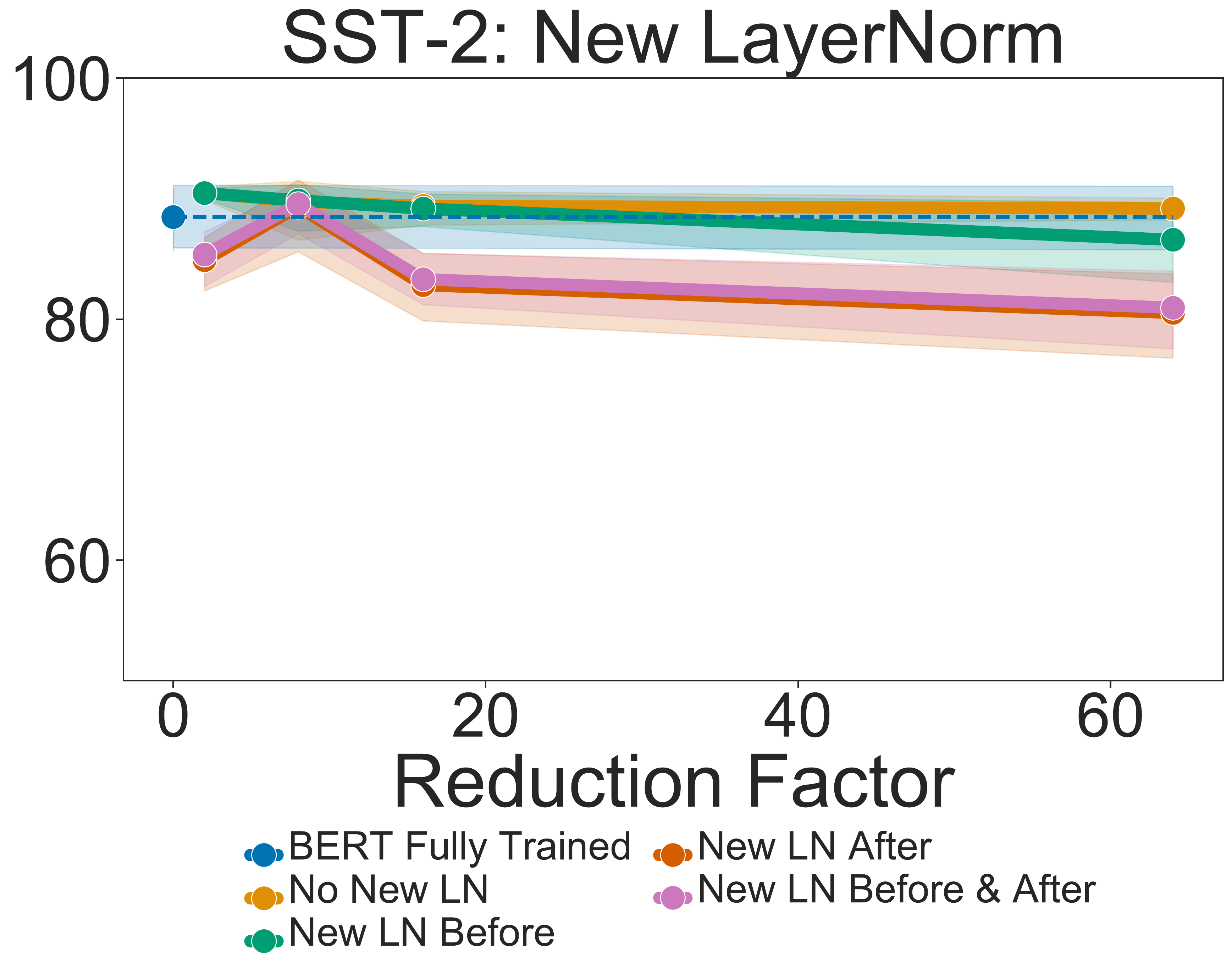}
\vphantom{\includegraphics[width=0.33\textwidth,valign=t]{img/SST_Adapters_2.pdf}}}
\caption{
Results of the grid search on the SST-2 dataset over the architecture settings illustrated on the left of Figure \ref{fig:Adapter_Architecture}. As we go from (a) to (c), the best performing setting is used for further search over other hyperparameters. We find that the best performing architecture is \textit{Top Adapter Only} with \textit{Pretrained LayerNorm Before \& After} including \textit{No New LayerNorm}. This Architecture is illustrated on the right of Figure \ref{fig:Adapter_Architecture}. 
}
\label{fig:SST_Adapter_search}
\end{figure*}

\begin{figure*}[htp]
\subfigure[Adapter Positions in Layer]
{\includegraphics[width=0.32\textwidth, valign=t ]{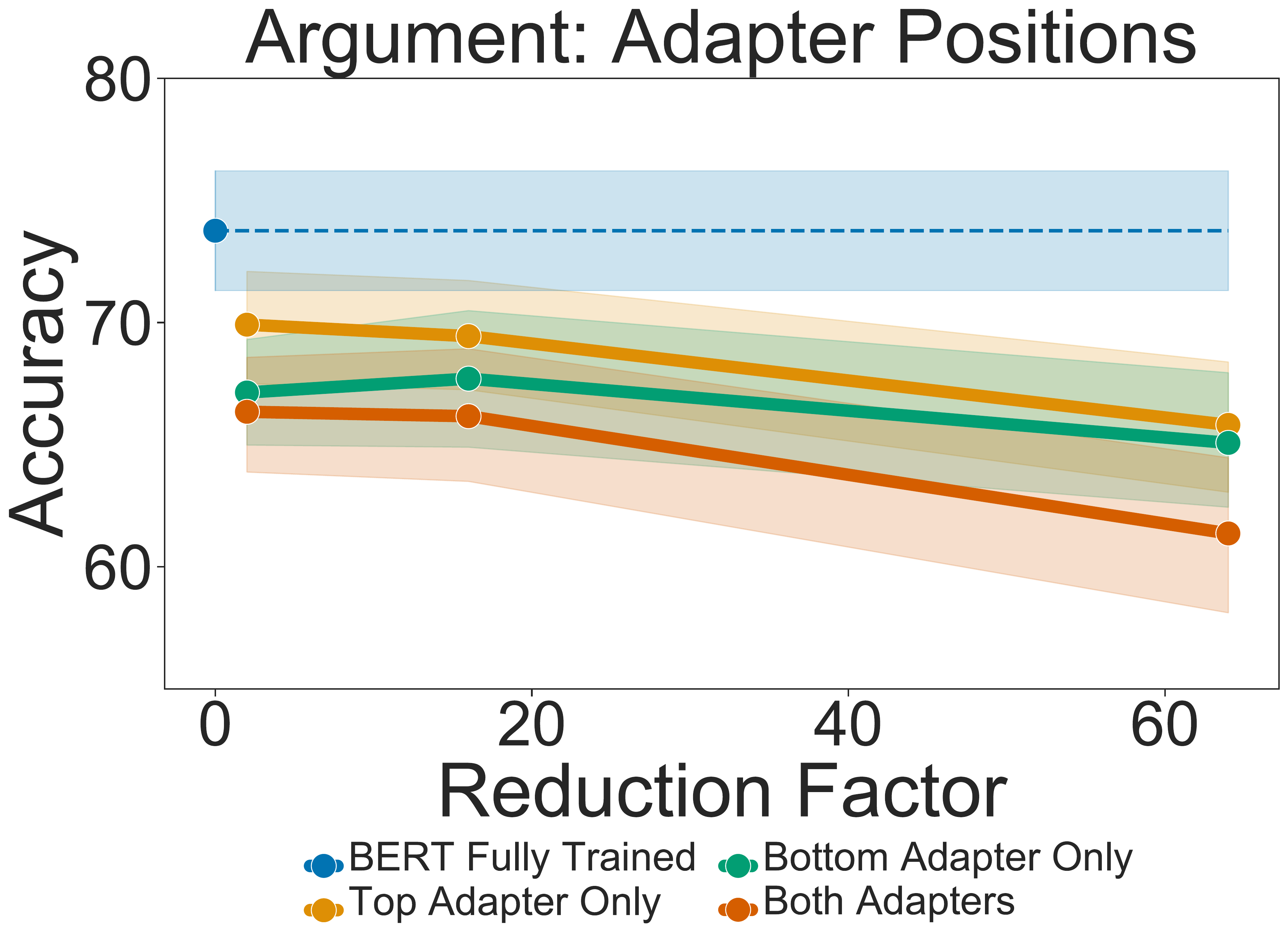}
\vphantom{\includegraphics[width=0.33\textwidth,valign=t]{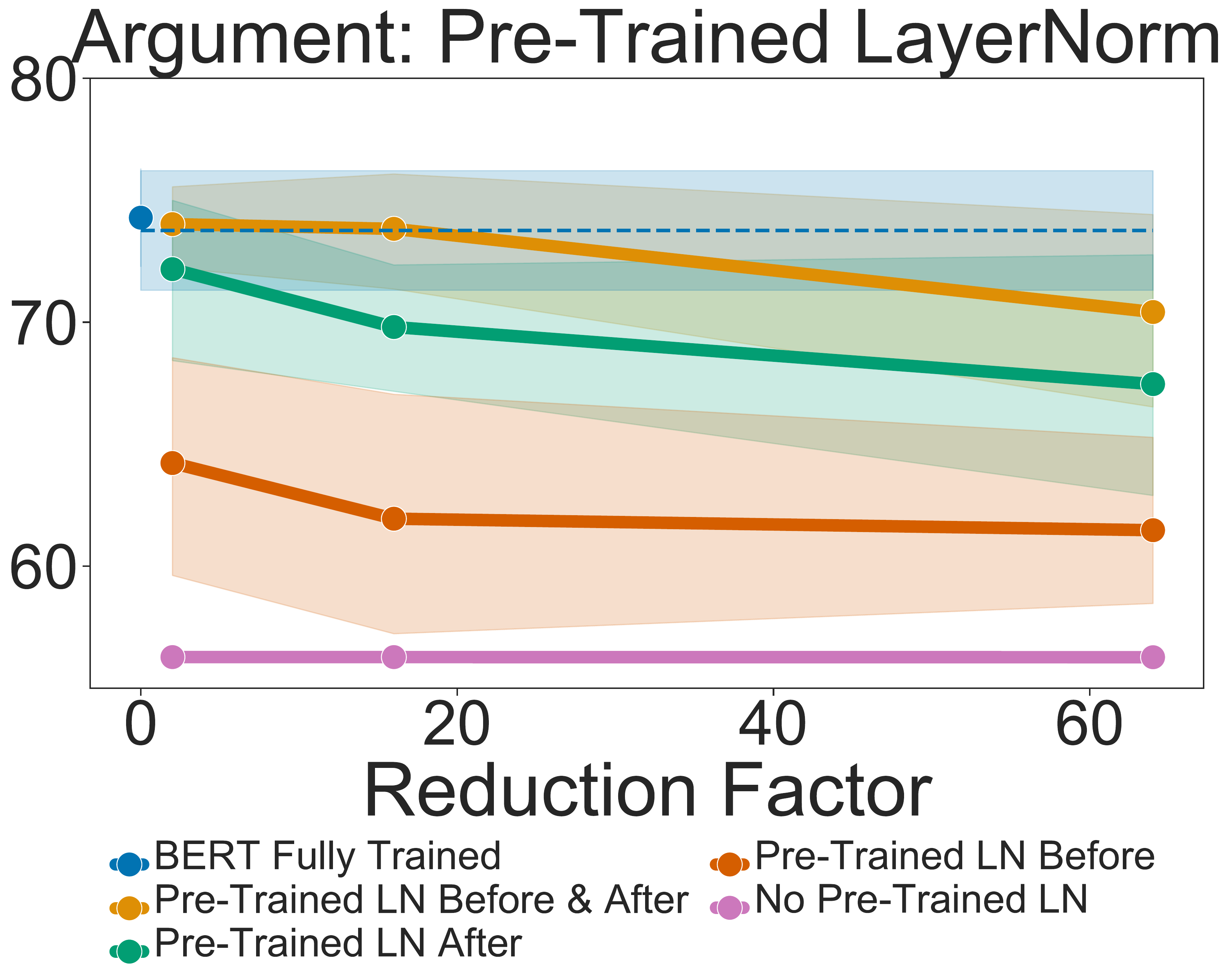}}
} \quad
\subfigure[Position of Pretrained LayerNorm ]{\includegraphics[width=0.3025\textwidth, valign=t ]{img/Argument_Adapters_2.pdf}
\vphantom{\includegraphics[width=0.33\textwidth,valign=t]{img/Argument_Adapters_2.pdf}}}
\quad
\subfigure[Position of newly trained LayerNorm   ]{\includegraphics[width=0.3025\textwidth, valign=t ]{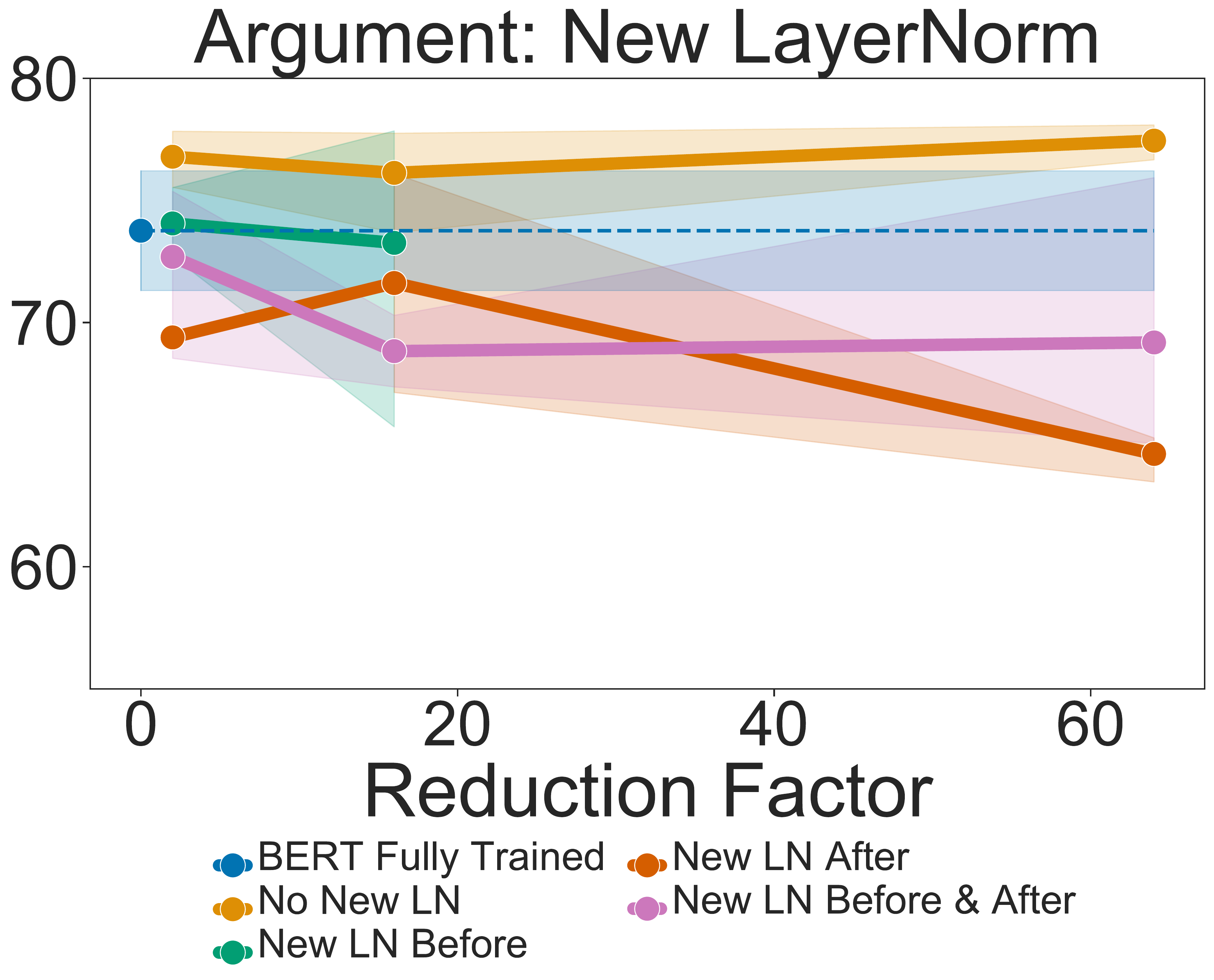}
\vphantom{\includegraphics[width=0.33\textwidth,valign=t]{img/Argument_Adapters_3.pdf}}}
\caption{
Results of the grid search on the Argument dataset over the architecture settings illustrated on the left of Figure \ref{fig:Adapter_Architecture}. As we go from (a) to (c), the best performing setting is used for further search over other hyperparameters. We find that the best performing architecture is \textit{Top Adapter Only} with \textit{Pretrained LayerNorm Before \& After} including \textit{No New LayerNorm}. This Architecture is illustrated on the right of Figure \ref{fig:Adapter_Architecture}. 
}
\label{fig:Argument_Adapter_search}
\end{figure*}

\begin{figure*}[htp]
\subfigure[Adapter Positions in Layer]
{\includegraphics[width=0.32\textwidth, valign=t ]{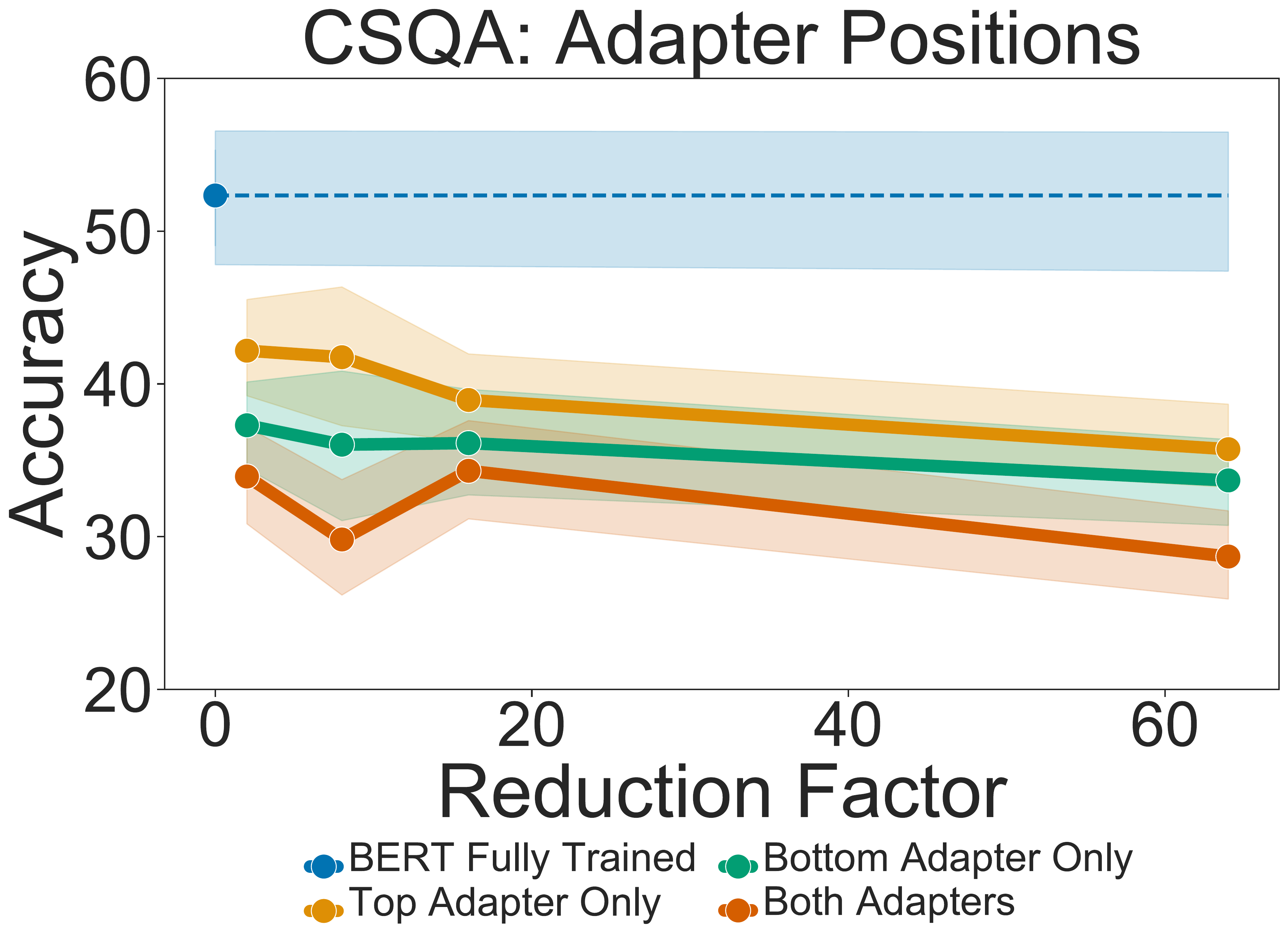}
\vphantom{\includegraphics[width=0.33\textwidth,valign=t]{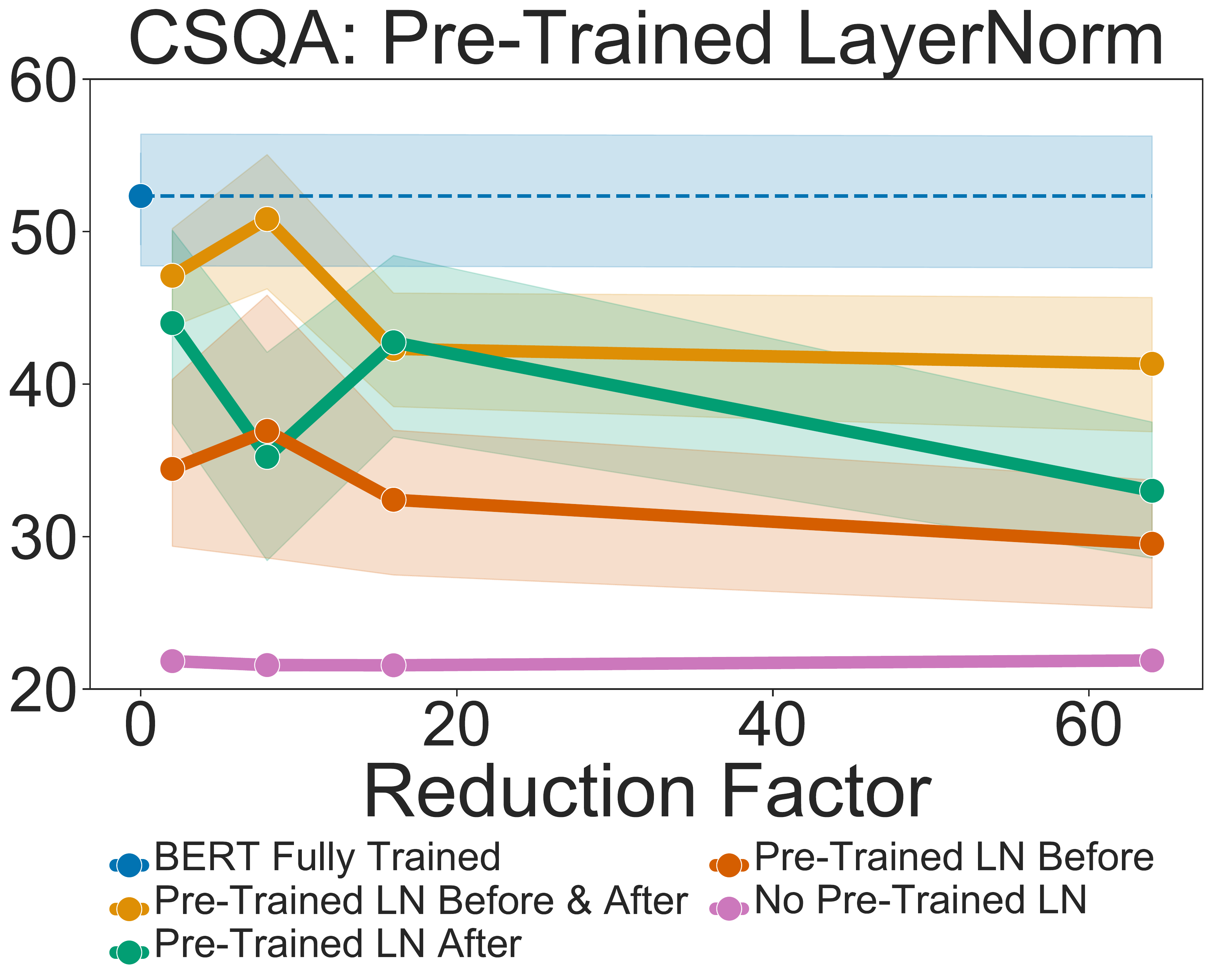}}
} \quad
\subfigure[Position of Pretrained LayerNorm ]{\includegraphics[width=0.3025\textwidth, valign=t ]{img/CSQA_Adapters_2.pdf}
\vphantom{\includegraphics[width=0.33\textwidth,valign=t]{img/CSQA_Adapters_2.pdf}}}
\quad
\subfigure[Position of newly trained LayerNorm   ]{\includegraphics[width=0.3025\textwidth, valign=t ]{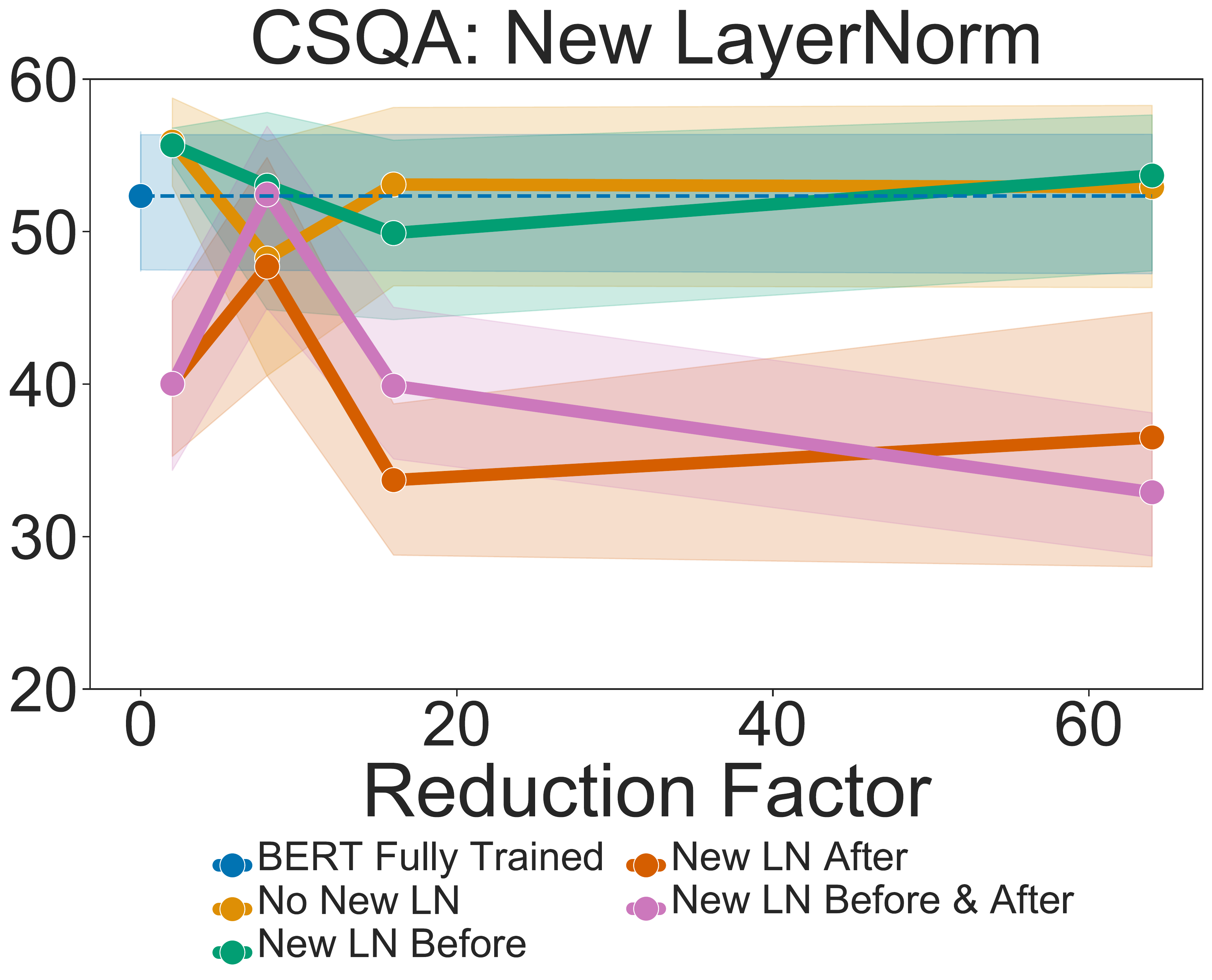}
\vphantom{\includegraphics[width=0.33\textwidth,valign=t]{img/CSQA_Adapters_3.pdf}}}
\caption{
Results of the grid search on the CSQA dataset over the architecture settings illustrated on the left of Figure \ref{fig:Adapter_Architecture}. As we go from (a) to (c), the best performing setting is used for further search over other hyperparameters. We find that the best performing architecture is \textit{Top Adapter Only} with \textit{Pretrained LayerNorm Before \& After} including \textit{No New LayerNorm}. This Architecture is illustrated on the right of Figure \ref{fig:Adapter_Architecture}. 
}
\label{fig:CSQA_Adapter_search}
\end{figure*}

As described in \S \ref{sec:adapterinpractice}, the placement of adapter parameters $\Phi$ within a pretrained model is non-trivial, and thus requires extensive experiments. In order to identify the best ST-A setting, we run an exhaustive architecture search on the hyperparameters --- including the position and number of adapters in each transformer layer, the position and number of pretrained or task dependent layer norms, the position of residual connections, the bottleneck reduction factors \{$2,8,16,64$\}, and the non linearity \{ReLU, LeakyReLU, Swish\} used within the adapter. We illustrate this in Figure~\ref{fig:Adapter_Architecture}. This grid search includes the settings introduced by \citet{houlsby2019parameter} and \citet{bapna2019simple}.
We perform this search on three diverse tasks\footnote{SST-2, Commonsense QA, and Argument.} and find that across all three tasks, the same setup obtains best results. 
We present our results on the SST-2, Argument, and CSQA datasets in Figures \ref{fig:SST_Adapter_search}, \ref{fig:Argument_Adapter_search}, and \ref{fig:CSQA_Adapter_search} respectively, at different granularity levels. 
We find that in contrast to \citet{houlsby2019parameter}, but in line with \citet{bapna2019simple}, a single adapter after the feed-forward layer outperforms other settings.  While we find that this setting performs on-par with that of \citet{houlsby2019parameter}, it requires only half the number of newly introduced adapters as compared to them, resulting in a more efficient setting in terms of number of operations.

For the single-task adapter setting, we thus perform all subsequent experiments with the
best architecture illustrated in Figure \ref{fig:Adapter_Architecture} on the right and a learning rate of $1e-4$. 
In order to reproduce the multi-task results in \citet{stickland2019bert} and build upon them, for experiments involving multi-task training, we adopt their architecture as described in \S \ref{sec:adapterinpractice}. 
 
\subsection{AdapterFusion Activations of all Layers}
\begin{figure*}[htp]
\centering
\includegraphics[width=1.0\linewidth]{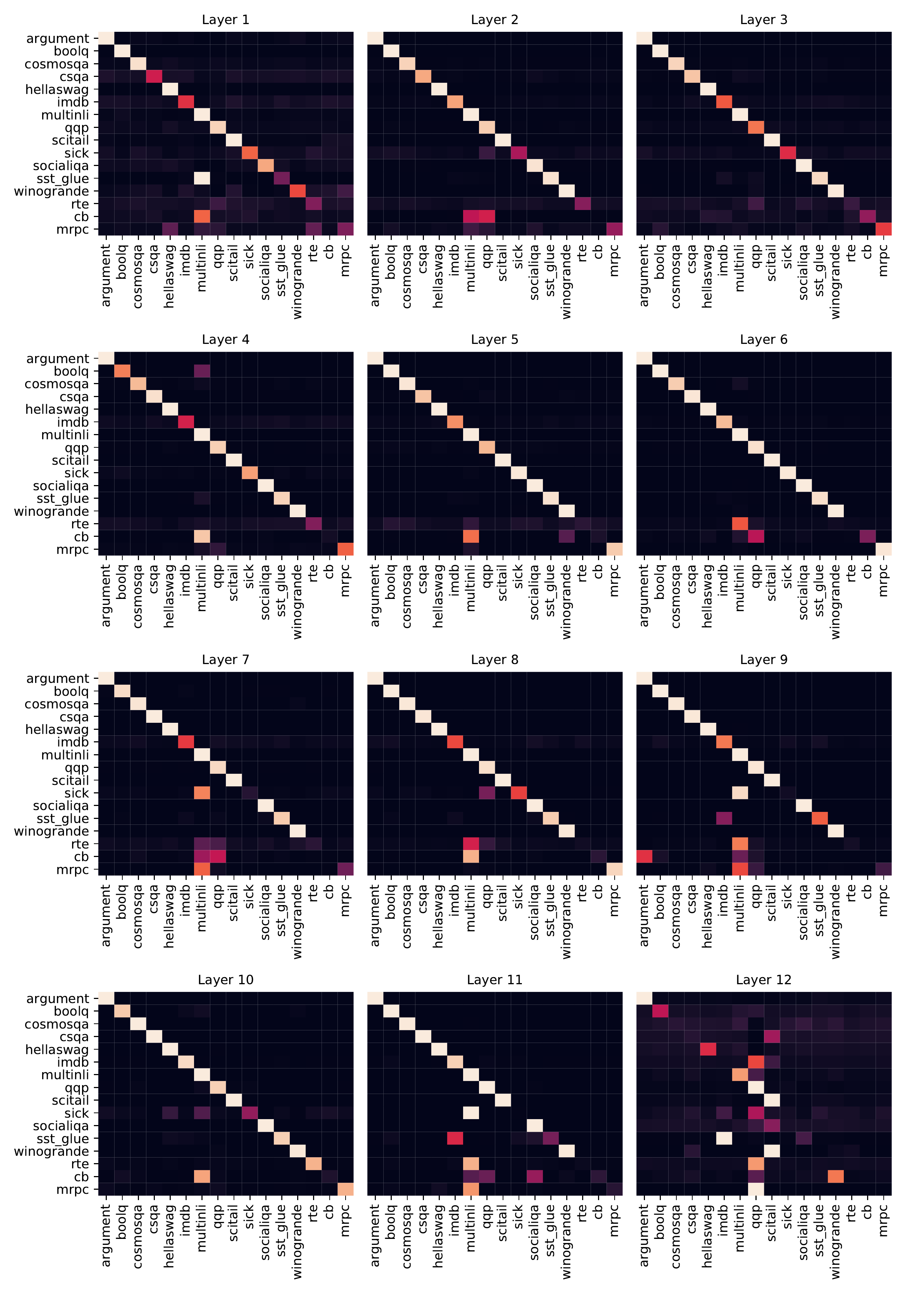}
\caption{AdapterFusion activations in the 12 BERT-base layers. Target tasks are presented in rows, whereas the set of adapters are displayed in columns. Black squares indicate that an adapter has not been activated, whereas white cells indicate full activation.  }
\label{fig:Heatmap_full}
\end{figure*}

We present the cross-product of activations of AdapterFusion of all layers for BERT-Base and ST-A$_{16}$  in Figure \ref{fig:Heatmap_full}, as an extension to Figure~\ref{fig:Heatmap_part_mt}.

\subsection{BERT-base ST-A with Reduction Factors \{$2, 16, 64$\}}
We present the ST-A results with different capacity leveraging BERT-base weights in Table~\ref{table:results_st_bert}. Reduction factors $2$, $16$, and $64$ amount to dense adapter dimensions $384$, $48$, and $12$ respectively.

\subsection{ST-A and Fusion with ST-A Results with RoBERTa-base}
In order to validate our findings of our best setup---ST-A---we re-evaluate our results leveraging RoBERTa-base weights. We present our results in Table~\ref{table:results_roberta}. Similar to our findigs with BERT-base, especially datasets with less data profit from AdapterFusion. We find that, in contrast to BERT-base, RoBERTa-base does not perform well with high capacity adapters with reduction factor 2.

\label{sec:appendix}

\begin{table*}[t!]
\centering
\footnotesize
\begin{tabular}{l|lll}
\toprule
\textbf{Dataset} &   \textbf{ST-A}$_2$ &   \textbf{ST-A}$_{16}$ & \textbf{ST-A}$_{64}$    \\  
\midrule 
MultiNLI	&	   84.60			&	84.32		 	&	84.08	 	  \\
QQP	        &	   90.57			&	90.59		 	&	89.73	 	  \\
SST     	&	   92.66		\stdintable{0.32}	&	91.85		\stdintable{0.41}	&	92.01	 	\stdintable{0.33} \\
Winogrande	&	   62.11		\stdintable{0.09}	&	61.09		\stdintable{0.11}	&	59.70	 	\stdintable{0.06} \\
IMDB    	&	   94.20		\stdintable{0.28}	&	93.85		\stdintable{0.07}	&	93.90	 	\stdintable{0.14} \\
HellaSwag	&	   39.45		\stdintable{0.20}	&	38.11		\stdintable{0.14}	&	38.28	 	\stdintable{0.37} \\
\hdashline
SocialIQA	&	   60.95		\stdintable{0.15}	&	62.41		\stdintable{0.11}	&	62.23	 	\stdintable{0.73} \\
CosmosQA	&	 	59.32		\stdintable{0.24}	&	60.01		\stdintable{0.02}	&	60.65	 	\stdintable{0.34} \\
SciTail	    &	   94.44		\stdintable{0.81}	&	93.90		\stdintable{0.16}	&	93.82	 	\stdintable{0.49} \\
Argument	&	 	76.83		\stdintable{0.21}	&	77.65		\stdintable{0.34}	&	77.64	 	\stdintable{0.56} \\
\hdashline
CSQA    	&	 	57.83		\stdintable{0.23}	&	58.91		\stdintable{0.57}	&	58.88	 	\stdintable{0.40} \\
BoolQ	    &	 	77.14		\stdintable{1.10}	&	75.66		\stdintable{1.25}	&	76.07	 	\stdintable{0.54} \\
MRPC    	&	 	86.13		\stdintable{1.59}	&	85.16		\stdintable{0.52}	&	85.58	 	\stdintable{0.32} \\
\hdashline
SICK	    &	   87.50		\stdintable{0.14}	&	86.20		\stdintable{0.00}	&	85.70	 	\stdintable{0.42} \\
RTE	        &	 	70.68		\stdintable{4.57}	&	71.04		\stdintable{1.62}	&	69.16	 	\stdintable{1.59} \\
CB	        &	 	87.85		\stdintable{2.94}	&	86.07		\stdintable{3.87}	&	84.28	 	\stdintable{4.79} \\

\midrule
\textbf{Mean}	&	76.39	  	& 76.05 				&  75.73	 	  	\\
\bottomrule
\end{tabular}
\caption{ Mean and standard deviation results (development sets) for each of the 16 datasets and reduction factors \{2, 16, 64\} for ST-A. Each model is initialized with BERT-base \cite{devlin2018bert} weights. The datasets are ordered by their respective training dataset size. Dashed horizontal lines separates datasizes \{$>40k, >10k, >5k$\} respectively.}
\label{table:results_st_bert}
\end{table*}

\begin{table*}[h]
\centering
\footnotesize
\begin{tabular}{l|ll:lll:l|l}
\toprule
\textbf{Dataset} &   \textbf{Head} &   \textbf{Full} & \textbf{ST-A$_{2}$}  & \textbf{ST-A$_{16}$}& \textbf{ST-A$_{64}$} & \textbf{F. w/ ST-A$_{16}$}  & \textbf{ST-A$_{16}^{\text{Houlsby}}$} \\  
\midrule 
MultiNLI	&	56.84	 	&	86.42	 	&	85.56	 	&	86.06	 	&	85.86	 	&	86.20	 	&	86.57	 	\\
QQP	&	71.40	 	&	91.07	 	&	90.88\stdintable{0.07}	&	90.27	 	&	89.39\stdintable{0.63}	&	90.28	 	&	90.66	 	\\
SST	&	81.86\stdintable{0.21}	&	94.29\stdintable{0.22}	&	93.71\stdintable{0.29}	&	93.80\stdintable{0.23}	&	93.35\stdintable{0.43}	&	93.67\stdintable{0.13}	&	94.17\stdintable{0.15}	\\
Winogrande	&	51.93	 	&	66.77	 	&	51.27\stdintable{0.78}	&	65.58\stdintable{0.53}	&	62.43	 	&	66.01\stdintable{0.47}	&	63.46\stdintable{6.38}	\\
IMDB	&	85.40	 	&	96.00	 	&	95.70	 	&	95.78\stdintable{0.13}	&	95.80	 	&	95.78\stdintable{0.19}	&	95.68\stdintable{0.26}	\\
HellaSwag	&	41.16	 	&	63.53	 	&	61.09	\stdintable{0.08}	&	61.57\stdintable{0.14}	&	61.18\stdintable{0.21}	&	61.52\stdintable{0.07}	&	61.21\stdintable{0.37}	\\
\hdashline
SocialIQA	&	46.87	 	&	69.44	 	&	69.24	 	&	70.14\stdintable{0.40}	&	70.21	 	&	70.13\stdintable{0.11}	&	70.78\stdintable{0.17}	\\
CosmosQA	&	41.88	\stdintable{0.29}	&	68.52\stdintable{0.49}	&	68.01\stdintable{0.94}	&	68.76\stdintable{0.53}	&	68.62\stdintable{0.55}	&	68.64\stdintable{0.04}	&	69.18\stdintable{0.34}	\\
SciTail	&	49.57	 	&	94.47	 	&	94.24	 	&	94.59\stdintable{0.64}	&	94.32	 	&	94.44\stdintable{0.09}	&	94.09\stdintable{0.39}	\\
Argument	&	66.22	\stdintable{0.62}	&	78.04\stdintable{0.42}	&	78.60\stdintable{0.34}	&	78.50\stdintable{0.45}	&	78.53\stdintable{0.59}	&	77.98\stdintable{0.24}	&	78.42\stdintable{0.44}	\\
\hdashline
CSQA	&	41.37\stdintable{0.34}	&	65.81\stdintable{0.59}	&	66.11\stdintable{0.60}	&	66.30\stdintable{0.38}	&	64.03\stdintable{0.27}	&	66.52\stdintable{0.18}	&	67.53\stdintable{0.70}	\\
BoolQ	&	62.17	 	&	81.89	 	&	80.86\stdintable{0.86}	&	80.83\stdintable{0.27}	&	80.17\stdintable{0.25}	&	80.86\stdintable{0.15}	&	81.11\stdintable{0.54}	\\
MRPC	&	68.38\stdintable{0.00}	&	89.11\stdintable{0.93}	&	89.11\stdintable{0.51}	&	88.72\stdintable{0.71}	&	87.10\stdintable{1.67}	&	89.65\stdintable{0.50}	&	89.17\stdintable{1.06}	\\
\hdashline
SICK	&	56.40	 	&	86.60	 	&	84.80	 	&	85.40\stdintable{0.32}	&	85.40	 	&	85.76\stdintable{0.26}	&	85.88\stdintable{0.46}	\\
RTE	&	55.81\stdintable{2.92}	&	72.34\stdintable{11.02}	&	61.80\stdintable{12.47}	&	75.30\stdintable{0.61}	&	73.86\stdintable{1.55}	&	78.79\stdintable{1.12}	&	78.56\stdintable{1.54}	\\
CB	&	59.64\stdintable{11.05}	&	90.00\stdintable{1.60}	&	87.14\stdintable{6.85}	&	89.28\stdintable{2.82}	&	81.07\stdintable{4.82}	&	92.86\stdintable{3.79}	&	89.64\stdintable{3.87}	\\

\midrule
\textbf{Mean} & 58.05 & 81.08 & 78.63 & 80.83 & 79.52 & \textbf{81.41} & 81.18 \\
\bottomrule
\end{tabular}
\caption{Mean and standard deviation results of models initialized with RoBERTa-base \cite{liu2019roberta} weights. Performances are measured on the development sets of the 16 datasets for the different architectural setups. The datasets are ordered by their respective training dataset size. Dashed horizontal lines separate datasizes \{$>40k, >10k, >5k$\} respectively. \textbf{Head} indicates training only a classification head on top of fixed RoBERTa weights. For \textbf{Full} training we fine-tune all weights of RoBERTa. Single-Task adapters (\textbf{ST-A}) is the training of independently trained adapters for each task, using the architecture illustrated in Figure \ref{fig:Adapter_Architecture}, indices \{2, 16, 64\} indicate the reduction factor. \textbf{Fusion w/ ST-A} show the results of AdapterFusion using the respective pretrained adapters. \textbf{ST-A$_{16}^{\text{Houlsby}}$} shows the results of ST-A with with architecture proposed by \citet{houlsby2019parameter}. }
\label{table:results_roberta}
\end{table*}

\end{document}